\documentclass[11pt,a4paper]{article}

\usepackage{amsmath,amssymb,bm,mathtools}
\usepackage[numbers,compress]{natbib}
\usepackage{graphicx}
\usepackage[font=small]{caption}

\newcommand\vc[1]{\mathbf{#1}}
\newcommand\vcs[1]{\bm{#1}}
\newcommand\mat[1]{\mathsf{#1}}
\newcommand\tp{\top}

\DeclareMathOperator\tr{tr}

\newcommand\expect[1]{{\left\langle#1\right\rangle}}

\renewcommand\d{\mathop{}\!\mathrm{d}}

\numberwithin{equation}{section}

\begin{document}
\bibliographystyle{unsrtnat}

\title{A Bayesian approach to type-specific conic fitting}
\author{M.J.Collett\\
Department of Physics, University of Auckland, New Zealand}
\date{}
\maketitle

\begin{abstract}
A perturbative approach is used to quantify 
the effect of noise in data points 
on fitted parameters in a general homogeneous linear model,
and the results applied to the case of conic sections.  
There is an optimal choice of normalisation that minimises bias, 
and iteration with the correct reweighting 
significantly improves statistical reliability.
By conditioning on an appropriate prior, 
an unbiased type-specific fit can be obtained.
Error estimates for the conic coefficients may also be used to obtain 
both bias corrections and confidence intervals for other curve parameters.
\end{abstract}

\section{Introduction}

Linear algebraic methods for fitting conic sections to noisy data
were introduced by Bookstein \cite{bookstein1979fitting}.
Subsequent work, mostly in the broad context of computer vision,
has discussed topics such as 
iterative refinement \cite{sampson1982fitting},
type-specific normalisation \cite{fitzgibbon1999direct, halir1998numerically}
and the correction of obvious normalisation bias \cite{harker2008direct} 
or curvature bias \cite{kanatani1994statistical,collett2014ellipse}.  
However, all of these papers confine themselves to making
point estimates of the set of conic coefficients, 
and do not quantify the expected statistical errors.

A recent series of papers by Kanatani and coworkers 
\cite{kanatani2011hyper, kanatani2011hyperls, kanatani2015hyper}
have used a perturbative treatment 
to predict the systematic errors
introduced into the fitting process by noisy data,
and hence to identify analytically 
(rather than merely numerically)
the choices of normalisation and weighting 
that minimise bias and statistical error.
These are forward statistical calculations, that is,
the noisy data is described by a sampling distribution around the true values;
the final result is an unbiased point estimate 
of the generic conic coefficients. 

In this paper I show that we can go significantly further 
by also using a Bayesian (inverse) statistical treatment,
that is, by considering a posterior distribution for the model parameters.

Knowing the posterior distribution has three distinct useful consequences.
Firstly, and most obviously, we can place confidence intervals on 
the estimated fit made from any particular data set;
since for the generic fit in the perturbative regime the distributions are Gaussian,
equivalent results could have been obtained from the sampling statistics. 
Secondly, and perhaps unexpectedly, being in possession of 
a distribution of fitted values rather than just a point estimate 
allows us to include 
in a systematic and unbiased fashion additional constraints, 
such as those required to make the fit type-specific. 
Finally, by propagating the estimated errors in the conic coefficients,
we may obtain both confidence intervals and bias corrections for estimates of other curve parameters (e.g.\ the location of the centre of an ellipse).

The method used has potential application to problems other than conic fitting
(e.g.\ camera resection \cite{collett2015self-normalising}),
so is initially developed for a generic homogeneous linear model,
before being applied (with numerical examples) to the conic case.
The initial sampling calculations here differ in detail from those of Kanatani \emph{et al.},
but yield similar initial point estimates.

\section{Errors in homogeneous model fitting}
\subsection{The model}
Consider a system described by 
an $M$-component homogeneous linear model of the form
\begin{equation}\label{eq:exactModel}
Z(\vc x) = \vc G^\tp \vc D(\vc x) = 0\ ,
\end{equation}
where $\vc G^\tp = \begin{pmatrix} g_1&\ldots&g_M \end{pmatrix}$
is the model vector and
\begin{equation}
\vc D^\tp(\vc x) = 
\begin{pmatrix} d_1(\vc x)&\ldots&d_M(\vc x) \end{pmatrix}
\end{equation} 
is the design vector for the $\Lambda$-dimensional data point $\vc x$.
The elements of $\vc D$ are known functions of $\vc x$ 
(e.g.\ powers of the components of $\vc x$),
and the model parameters $g_m$ are to be fitted from the data.
I confine myself here to the case of scalar $Z$, 
but a multicomponent algebraic error may be treated in similar fashion.
To avoid the trivial solution $\vc G=\vc 0$, 
we impose a normalisation constraint of the form
\begin{equation}\label{eq:normalise}
\vc G^\tp\mat C\vc G = 1\ .
\end{equation}
The $M\times M$ constraint matrix $\mat C$ 
may be constant or depend on the data;
its rank $R$ determines the dimension of the solution space.

An observed data point $\widehat{\vc x}$ 
includes measurement error and other sources of noise, 
\begin{equation}
\widehat{\vc x} = \vc x + \Delta \vc x\ ,
\end{equation}
with the consequence that 
\eqref{eq:exactModel} is not exactly satisfied.
Given a set of $N$ such data points that overdetermines the solution,
we therefore look for the model vector ${\vc G}$ that minimises
\begin{align}\label{eq:scatter}
E &= \sum_{i=1}^N w_iZ^2(\widehat{\vc x}_i) \nonumber\\
&= \sum_{i} w_i{\vc G}^\tp \vc D(\widehat{\vc x}_i)\vc D^\tp(\widehat{\vc x}_i){\vc G}
= {\vc G}^\tp{\mat S}{\vc G}\ ,
\end{align}
where
\begin{equation}
{\mat S} = \sum_{i=1}^N w_i\vc D(\widehat{\vc x}_i)\vc D(\widehat{\vc x}_i)^\tp
\end{equation}
is the $M\times M$ scatter matrix, 
and $w_i$ are statistical weights to be chosen later.
We still require normalisation, so that we in fact minimise the ratio
\begin{equation}
\frac{{\vc G}^\tp{\mat S}{\vc G}}{{\vc G}^\tp{\mat C}{\vc G}}\ ,
\end{equation}
which we can do by solving the generalised eigenvalue problem
\begin{equation}
{\mat S}{\vc G} = \lambda{\mat C}{\vc G}\ .
\label{eq:fullEigen}\end{equation}

We choose a basis for our model such that 
the constraint matrix may be partitioned into 
an $R\times R$ upper-left corner $\widetilde{\mat C}$ of full rank 
and zeros everywhere else,
\begin{equation}
\mat C =
\begin{pmatrix}
\widetilde{\mat C} & \mat 0_{RK} \\
\mat 0_{KR} & \mat 0_{KK}
\end{pmatrix}\ ,
\label{eq:partitionC}\end{equation}
where $K=M-R$.
We then partition $\mat S$ and $\vc G$ similarly:
\begin{equation}
\mat S = \begin{pmatrix}
\mat S_{11} & \mat S_{12} \\
\mat S_{21} & \mat S_{22}
\end{pmatrix}
\ ;\quad\vc G = 
\begin{pmatrix}\widetilde{\vc G}\\\widetilde{\vc H}\end{pmatrix}\ .
\end{equation}
The $K\times K$ matrix $\mat S_{22}$ is positive definite;
the $R\times R$ matrix $\mat S_{11}$ has a zero eigenvalue 
in the absence of noise, but is positive definite for real data.
The eigenvalue problem \eqref{eq:fullEigen} is now equivalent to the reduced version
\begin{equation}\label{eq:reducedEigen}
\widetilde{\mat S} \widetilde{\vc G} = \lambda\widetilde{\mat C}\widetilde{\vc G}\ ,
\end{equation}
where the reduced scatter matrix is the Schur complement
\begin{equation}
\widetilde{\mat S}=\mat S_{11}- \mat S_{12}\mat S_{22}^{-1}\mat S_{21},
\end{equation}
and the full eigenvector can be reconstructed using 
\begin{equation}
\widetilde{\vc H} = -\mat S_{22}^{-1}\mat S_{21}\widetilde{\vc G}\ .
\end{equation}
$\widetilde{\mat S}$ may be expressed in diagonalised  form as
\begin{equation}
\widetilde{\mat S} = 
\sum_{m=0}^{R-1}\widetilde{\mat C}\widetilde{\vc G}_m{\lambda_m}\widetilde{\vc G}_m^\tp\widetilde{\mat C} \ ,
\end{equation}
and hence $\mat S$ as
\begin{equation}
\mat S = \sum_{m}\mat C{\vc G}_m{\lambda_m}{\vc G}_m^\tp\mat C + 
\begin{pmatrix} \mat S_{12} \\ \mat S_{22} \end{pmatrix}\mat S_{22}^{-1}
  \begin{pmatrix} \mat S_{21} & \mat S_{22} \end{pmatrix}\ .
\end{equation}

\subsection{Perturbative treatment of errors}

If there is no noise in the measured values, 
then the smallest eigenvalue $\lambda_0$ 
in \eqref{eq:fullEigen} or \eqref{eq:reducedEigen} is zero,
and the corresponding eigenvector ${\vc G}_0$ is 
the exact solution to the original problem.  
With the inclusion of noise, the eigenvector ${\vc G}_0$ will
in general differ by both systematic bias and random error from the true solution.

Consider the effect on the $n$th eigenvalue $\lambda_n$ 
and its associated reduced eigenvector $\widetilde{\vc G}_n$
of a small change $\Delta\widetilde{\mat S}$ in $\widetilde{\mat S}$ 
(and any associated change in $\widetilde{\mat C}$).
\eqref{eq:reducedEigen} becomes 
\begin{align}\label{eq:perturb}
\left(\widetilde{\mat S}+\Delta \widetilde{\mat S}\right)\left(\widetilde{\vc G}_n+\Delta \widetilde{\vc G}_n\right)
= \left(\lambda_n+\Delta \lambda_n\right)
\left(\widetilde{\mat C}+\Delta \widetilde{\mat C}\right)\left(\widetilde{\vc G}_n+\Delta \widetilde{\vc G}_n\right)
\ .
\end{align}
Expanding and keeping only first-order changes gives
\begin{equation}
\Delta\widetilde{\mat S}\widetilde{\vc G}_n+\widetilde{\mat S}\Delta \widetilde{\vc G}_n 
\simeq \Delta\lambda_n\widetilde{\mat C}\widetilde{\vc G}_n
+\lambda_n\Delta\widetilde{\mat C}\widetilde{\vc G}_n
+\lambda_n\widetilde{\mat C}\Delta\widetilde{\vc G}_n
\ .
\end{equation}
Multiplying on the left by $\widetilde{\vc G}_m^\tp$, 
and using ${\widetilde{\vc G}_m^\tp\widetilde{\mat S}}={\lambda_m\widetilde{\vc G}_m^\tp\widetilde{\mat C}}$
and the orthonormality of the eigenvectors,
\begin{equation}\label{eq:orthonormal}
\widetilde{\vc G}_m^\tp\widetilde{\mat C}\widetilde{\vc G}_n=\delta_{mn}\ ,
\end{equation}
gives us the first-order results
\begin{subequations}
\begin{align}
\widetilde{\vc G}_n^\tp\left(\Delta\widetilde{\mat S}
-\lambda_n\Delta\widetilde{\mat C}\right)\widetilde{\vc G}_n&=
\Delta\lambda_n\ ;\label{eq:perturbVal}\\
\widetilde{\vc G}_m^\tp\left(\Delta\widetilde{\mat S}
-\lambda_n\Delta\widetilde{\mat C}\right)\widetilde{\vc G}_n
&=(\lambda_n-\lambda_m)\widetilde{\vc G}_m^\tp\widetilde{\mat C}\Delta\widetilde{\vc G}_n
\ ,\ m\ne n\ .
\end{align}
\end{subequations}
Thus we have an expansion of $\Delta\widetilde{\vc G}_n$ 
over a basis of the unperturbed eigenvectors in the form
\begin{equation}\label{eq:perturbVec}
\Delta\widetilde{\vc G}_n  = 
\sum_{m\ne n} \frac{\widetilde{\vc G}_m\widetilde{\vc G}_m^\tp}{\lambda_n-\lambda_m}
\left(\Delta\widetilde{\mat S}-\lambda_n\Delta\widetilde{\mat C}\right)\widetilde{\vc G}_n\ .
\end{equation}
Apart from the $\Delta\widetilde{\mat C}$ terms 
(which are zero for fixed constraints and of little consequence otherwise), 
\eqref{eq:perturbVal} and \eqref{eq:perturbVec} are precisely 
the results of standard first-order perturbation theory 
familiar from any quantum mechanics textbook.

The starting point for our perturbation is the noise-free case, 
for which as previously noted $\lambda_0=0$,
and we are primarily interested in the correction to ${\vc G}_0$,
giving the simplified result that
\begin{equation}\label{eq:perturbReduce0}
\Delta\widetilde{\vc G}_0 = 
-\widetilde{\mat Y}_0\Delta\widetilde{\mat S}\widetilde{\vc G}_0\ ,
\end{equation}
where
\begin{equation}
\widetilde{\mat Y}_0 =
\sum_{m\ne 0} \frac{\widetilde{\vc G}_m\widetilde{\vc G}_m^\tp}{\lambda_m}
\end{equation}
is a generalised inverse of the noise-free reduced scatter matrix.
It is not the Moore-Penrose pseudoinverse, but is closely related to it; factorising the normalisation matrix as 
$\widetilde{\mat C} = \mat J \mat J^\tp$
we find that
\begin{equation}
\widetilde{\mat Y}_0 = \mat J^{-\tp}\left(\mat J^{-1} 
  \widetilde{\mat S} \mat J^{-\tp}\right)^+\mat J^{-1}\ ,
\label{eq:pseudoinverse}\end{equation}
where superscript $+$ denotes the pseudoinverse.

To relate the change in $\widetilde{\mat S}$ to 
underlying changes in the data points being fitted,
it will be more convenient to return to the unreduced representation.
To first order,
\begin{align}
\Delta\widetilde{\mat S} = \Delta\mat S_{11}
- \mat S_{12}\mat S_{22}^{-1}\Delta\mat S_{21} 
- \Delta\mat S_{12}\mat S_{22}^{-1}\mat S_{21}
+ \mat S_{12}\mat S_{22}^{-1}\Delta\mat S_{22}\mat S_{22}^{-1}\mat S_{21}\ ,
\end{align}
giving
\begin{align}
\widetilde{\vc G}_m^\tp\Delta\widetilde{\mat S}\widetilde{\vc G}_n 
&= \widetilde{\vc G}_m^\tp\Delta\mat S_{11}\widetilde{\vc G}_n 
+ \widetilde{\vc H}_m^\tp\Delta\mat S_{21}\widetilde{\vc G}_n
+ \widetilde{\vc G}_m^\tp\Delta\mat S_{12}\widetilde{\vc H}_n 
 + \widetilde{\vc H}_m^\tp\Delta\mat S_{22}\widetilde{\vc H}_n\nonumber\\
&= {\vc G}_m^\tp\Delta{\mat S}{\vc G}_n\ ;
\end{align}
a corresponding result holds trivially for $\Delta\widetilde{\mat C}$.  So now
\begin{align}
\Delta{\vc G}_n &= \begin{pmatrix}\Delta\widetilde{\vc G}_n\\
\Delta\widetilde{\vc H}_n\end{pmatrix} 
= \begin{pmatrix}\Delta\widetilde{\vc G}_n\\
\Delta\left(-\mat S_{22}^{-1}\mat S_{21}\widetilde{\vc G}_n\right)\end{pmatrix}\nonumber\\
&= \begin{pmatrix}\Delta\widetilde{\vc G}_n\\
\big(-\mat S_{22}^{-1}\mat S_{21}\Delta\widetilde{\vc G}_n 
-\mat S_{22}^{-1}\Delta\mat S_{21}\widetilde{\vc G}_n \\
\qquad\qquad+ \mat S_{22}^{-1}\Delta\mat S_{22}
\mat S_{22}^{-1}\mat S_{21}\widetilde{\vc G}_n \big)
\end{pmatrix}
\nonumber\\
&=\begin{pmatrix}\Delta\widetilde{\vc G}_n\\
-\mat S_{22}^{-1}\mat S_{21}\Delta\widetilde{\vc G}_n\end{pmatrix}
-\begin{pmatrix} \mat 0_{RR} & \mat 0_{RK} \\ 
\mat 0_{KR} & \mat S_{22}^{-1} \end{pmatrix}\Delta\mat S
\begin{pmatrix}\widetilde{\vc G}_n\\
-\mat S_{22}^{-1}\mat S_{21}\widetilde{\vc G}_n\end{pmatrix}\nonumber\\
&= -\mat Y_n\left(\Delta{\mat S}-\lambda_n\Delta{\mat C}\right){\vc G}_n\ ,
\end{align}
where
\begin{equation}
\mat Y_n = \sum_{m\ne n} \frac{{\vc G}_m{\vc G}_m^\tp}{\lambda_m-\lambda_n} 
+ \begin{pmatrix} \mat 0_{RR} & \mat 0_{RK} \\ 
\mat 0_{KR} & \mat S_{22}^{-1} \end{pmatrix}\ .
\label{eq:fullY}\end{equation}
Again specialising to the change in ${\vc G}_0$ from the noise-free case gives
\begin{equation}\label{eq:perturbVec0}
\Delta{\vc G}_0 = -\mat Y_0\Delta{\mat S}{\vc G}_0\ .
\end{equation}

\subsection{Dependence on measurement noise}

Since the independent variables are random, 
a consistent calculation of the first-order change in $\mat S$
requires us to consider not only the first-order dependence on each $\Delta\vc x_i$, 
but also the expected value (i.e.\ the deterministic part) of the second-order dependence.

The former gives
\begin{equation}
\left(\Delta \mat S\right)_1 = 
\sum_{i=1}^N\sum_{\mu=1}^\Lambda w_i\Delta x_{i\mu}
\left(\vc D_{i,\mu}\vc D_i^\tp + \vc D_i\vc D_{i,\mu}^\tp\right)
\end{equation}
and the latter
\begin{align}
\left(\Delta \mat S\right)_2 = 
\frac12\sum_{i\mu\nu} w_i\expect{\Delta x_{i\mu}\Delta x_{i\nu}}
\left(\vc D_{i,\mu}\vc D_{i,\nu}^\tp 
 + \vc D_{i,\nu}\vc D_{i,\mu}^\tp + \vc D_i\vc D_{i,\mu\nu}^\tp 
+ \vc D_{i,\mu\nu}\vc D_i^\tp\right)
\end{align}
If the noise is homogeneous and isotropic, 
with $\expect{\Delta x_{i\mu}\Delta x_{i\nu}}=\sigma^2\delta_{\mu\nu}$,
this reduces to
\begin{align}
\left(\Delta \mat S\right)_2 = 
\sigma^2{\mat C}_\text{N} +\sigma^2 ({\mat K}+ {\mat K}^\tp)\ ,
\end{align}
where
\begin{subequations}
\begin{align}
{\mat C}_\text{N} &= \sum_{i\mu} w_i\vc D_{i,\mu}\vc D_{i,\mu}^\tp\ ; \label{eq:defCN}\\
\quad\mat K &= \tfrac12\sum_{i\mu} w_i\vc D_i\vc D_{i,\mu\mu}^\tp 
= \tfrac12\sum_{i} w_i\vc D_i\nabla^2\vc D_{i}^\tp\ .
\end{align}
\end{subequations}

Substituting back into \eqref{eq:perturbVal} with $n=0$, 
most terms vanish, leaving only the term in ${\mat C}_\text{N}$,
\begin{equation}\label{eq:noiseVal}
\Delta\lambda_0 = \sigma^2{\vc G}_0^\tp{\mat C}_\text{N}{\vc G}_0\ ;
\end{equation}
if we choose our constraint matrix to be ${\mat C}_\text{N}$ 
we have by \eqref{eq:normalise} the simple result 
that ${\Delta\lambda_0 = \sigma^2}$.

Similarly substituting back into \eqref{eq:perturbVec0}, 
some terms again vanish using the fact that 
the unperturbed vectors satisfy $\vc D_i^\tp \vc G_0= 0$.
However, we are still left with three distinct contributions to 
the error in the fitted parameters: 
\begin{align}
\Delta{\vc G}_0 
= -\mat Y_0\sum_{i\mu} w_i\Delta x_{i\mu}\vc D_i\vc D_{i,\mu}^\tp{\vc G}_0
-\sigma^2\mat Y_0{\mat C}_\text{N}{\vc G}_0 
-\sigma^2\mat Y_0{\mat K}{\vc G}_0\ .
\label{eq:fitErrors}\end{align}
The first term is the zero-mean random error, 
which will be our main interest for the remainder of this paper; 
the second is a normalisation bias, 
which vanishes by \eqref{eq:fullY} and \eqref{eq:orthonormal} 
if we choose our constraint matrix to be ${\mat C}_\text{N}$; 
the third is a curvature bias \cite{kanatani1994statistical}, 
which arises from the fact that $\expect{Z(\widehat{\vc x}_i)}\ne0$.
We expect the random error to scale as $\sigma/\sqrt{N}$, 
and the bias terms to scale as $\sigma^2/R$, 
where $R$ is the minimum radius of curvature of the curve or surface to be fitted:
for small noise ($\sigma\ll R$) and few data points the random error will dominate;
for larger noise (still with $\sigma< R$) and many points the biases are more important.

To avoid the curvature bias, we need to correct \eqref{eq:scatter} to
\begin{align}\label{eq:scatterDebias}
E &= \sum_i w_i\left(Z(\widehat{\vc x}_i)-\expect{Z(\widehat{\vc x}_i)}\right)^2 \nonumber\\
&= \sum_i w_i{\vc G}^\tp\left(\vc D(\widehat{\vc x}_i)-\expect{\vc D(\widehat{\vc x}_i)}\right)
\left(\vc D(\widehat{\vc x}_i)-\expect{\vc D(\widehat{\vc x}_i)}\right)^\tp{\vc G},
\end{align}
where with the assumptions already made
\begin{equation}
\expect{\vc D(\widehat{\vc x}_i)} = \tfrac12\sigma^2\nabla^2\vc D({\vc x}_i) 
\simeq \tfrac12\sigma^2\nabla^2\vc D(\widehat{\vc x}_i)\ .
\end{equation}
If, as will commonly but not invariably be the case, 
the components of $\nabla^2\vc D$ can be expressed in terms of those of $\vc D$ as
\begin{equation}
\nabla^2\vc D = 2\mat L \vc D
\label{eq:linearLaplace}\end{equation}
for some constant matrix $\mat L$, \eqref{eq:scatterDebias} simplifies to
\begin{align}
E &= \sum_i w_i{\vc G}^\tp(1-\sigma^2\mat L)\vc D(\widehat{\vc x}_i)\vc D^\tp(\widehat{\vc x}_i)(1-\sigma^2\mat L^\tp){\vc G}\nonumber\\
&= {\vc G}^\tp(1-\sigma^2\mat L)\mat S (1-\sigma^2\mat L^\tp){\vc G}\ .
\end{align}
This in turn implies that the generalised eigenvector of $\mat S$ denoted $\vc G_0$ 
is not in fact an estimator of $\vc G$, 
but rather of $(1-\sigma^2\mat L^\tp){\vc G}$.
The corrected estimator of $\vc G$ is accordingly 
\begin{equation}
\vc G = (1-\sigma^2\mat L^\tp)^{-1}{\vc G_0}\simeq(1+\sigma^2\mat L^\tp){\vc G_0}\ ,
\label{eq:curvCorrect}\end{equation}
where we can obtain a value for $\sigma^2$ with the aid of \eqref{eq:noiseVal}.
Under the same conditions we have that $\mat K = \mat S\mat L^\tp$, 
and hence that the curvature bias term is 
$-\sigma^2\mat Y_0{\mat S}\mat L^\tp{\vc G}_0$.
Combining the two, we have a corrected curvature bias
\begin{equation}
(\Delta{\vc G})_{\mat K} = \sigma^2\left(\mat I_M-\mat Y_0{\mat S}\right)\mat L^\tp{\vc G}_0\ .
\label{eq:residCurv}\end{equation}
$\mat I_M-\mat Y_0{\mat S}$ projects onto ${\vc G}_0$, 
so any remaining effect is purely a rescaling of the entire model vector,
and hence irrelevant for a homogeneous model.

Clearly $\mat C_\text{N}$ as given by \eqref{eq:defCN} is to first order
an optimal choice of constraint matrix: 
it avoids normalisation bias, and directly gives 
an estimate of $\sigma^2$ and hence of the required curvature bias correction. 
This normalisation is equivalent 
both to Taubin's approximate mean-square distance \cite{taubin1991estimation}
and to the method advocated by Harker and O'Leary \cite{harker2006direct}.
In those cases, however, 
it appears as the result of an average over optimal weights---%
in the former case, those best approximating geometric distance, 
and in the latter, those giving a statistically ideal least-squares fit.
By contrast, we have not yet considered the question of choice of weights 
(though we are about to do so),
only systematic bias.

A plausible alternative choice of constraint matrix,
equivalent to that made in the conic case by \cite{kanatani2011hyper},  
is $\mat C = \mat C_\text{N}+{\mat K}+ {\mat K}^\tp$,
which would directly eliminate both bias terms 
and still leave ${\Delta\lambda_0 = \sigma^2}$ to first order.
The downside is that this will rarely be in the desired form \eqref{eq:partitionC},
requiring additional computation to change to a basis in which it is.
If the reduction \eqref{eq:linearLaplace} 
(and hence the simple curvature correction \eqref{eq:curvCorrect}) is not available, 
this may nevertheless be the optimal choice.
However, if the reduction \eqref{eq:linearLaplace} does hold,
then this choice gives to first order the same results as
choosing $\mat C = \mat C_\text{N}$ and applying the curvature correction, 
since \eqref{eq:fullEigen} becomes
\begin{align}
{\mat S}{\vc G} &= \lambda\left(\mat C_\text{N}+{\mat K}+ {\mat K}^\tp\right){\vc G}
= \lambda\left(\mat C_\text{N}+ {\mat S}{\mat L}^\tp+{\mat L}{\mat S}\right){\vc G}
\ ,
\end{align}
and hence
\begin{align}
(\mat I -\lambda{\mat L}) {\mat S}(\mat I -\lambda{\mat L}^\tp){\vc G} &= \lambda\mat C_\text{N}{\vc G}\ .
\end{align}

From the first term in \eqref{eq:fitErrors}, the covariance matrix of the vector of coefficients is
\begin{equation}
\mat V_0 = \expect{\Delta{\vc G}_0,\Delta{\vc G}_0^\tp} 
= \sigma^2\mat Y_0 \sum_{i\mu} w_i^2\vc D_i(\vc D_{i,\mu}^\tp{\vc G}_0)^2\vc D_i^\tp\mat Y_0\ .
\label{eq:covariance}\end{equation}
By differentiating with respect to $w_i$ we see that 
the optimal choice of weighting 
(i.e.\ the one that minimises the variances of the fitted parameters) is 
\begin{equation}
w_i^{-1} = N \sum_{\mu} (\vc D_{i,\mu}^\tp{\vc G}_0)^2\ ,
\label{eq:optimalWeight}\end{equation}
but to use this we must already have an approximate value for the model vector.
It thus requires an iterative approach, 
in which we obtain an initial estimate for ${\vc G}_0$ using constant weights
and then refine it with the improved weights.
Iteration in this fashion was suggested by Sampson \cite{sampson1982fitting},  
but Sampson's weighting, based on the gradient at the measured data values, 
introduces bias into the fit \cite{collett2014ellipse}
and does not converge reliably for large noise \cite{gander1994least-squares}.
This weighting bias is similar in magnitude to the curvature bias, 
but its effects are more widespread, not confined to regions of high curvature.
To avoid it, the gradient of the design vector $\vc D_{i,\mu}$ 
appearing in \eqref{eq:optimalWeight} must be evaluated 
not at the measured point $\widehat{\vc x}_i$ 
but instead at a nearby point consistent with (the current estimate of) the model;
the best method of finding this point depends on details of the model being fitted.   
(Note that this is a less significant issue 
when calculating the constraint matrix ${\mat C}_\text{N}$ from \eqref{eq:defCN};
in that case we can to leading order safely evaluate $\vc D_{i,\mu}$ at the measured points,
since the summation over $i$ averages over individual deviations.)
With the optimal weighting \eqref{eq:optimalWeight} 
the quantity $E$ that the fit minimises is 
$\sigma^{-2}$ times the conventional $\chi^2$ statistic for $Z(\widehat{\vc x})$;
to leading order in the size of the measurement noise, 
this is also equal to the geometric mean-square error.
Although it is important that the reweighting process not introduce bias,
it does not otherwise need to be very precise;
the first reweighting may give a significant reduction in the random error,
but there is typically little further gain from repeated iterations.
It should be stressed (since it has not always been clear in the existing literature) 
that the purpose of reweighting is to improve the precision of the fit, 
not its accuracy; it is not an effective technique for the reduction or removal of bias.

With the weighting \eqref{eq:optimalWeight} (or an adequate approximation to it), 
the resulting covariance matrix is 
\begin{equation}
\mat V_0 = \frac{\sigma^2}{N}\mat Y_0 \sum_i w_i\vc D_i\vc D_i^\tp\mat Y_0 
= \frac{\sigma^2}{N}\mat Y_0 \mat S\mat Y_0 = \frac{\sigma^2}{N}\mat Y_0 \ ;
\label{eq:optimalCovariance}\end{equation}
for $K=1$ we can equivalently write
\begin{equation}
\Delta{\vc G}_0 = \eta_0 \frac{\sigma}{\sqrt{NS_{22}}} {\vc H}_0 + \sum_{m\ne 0} \eta_m \frac{\sigma}{\sqrt{N\lambda_m}}{\vc G}_m \ ,
\end{equation}
where $\eta_m$ are independent unit-variance random variables and 
${\vc H}_0^\tp = \begin{pmatrix} {\vc 0}_R^\tp & 1 \end{pmatrix}$.
Alternatively, with a Gaussian model for the errors, 
the sampling distribution for $\Delta{\vc G}_0$ 
(or equivalently for $\widehat{\vc G}_0 = \vc G_0 +\Delta{\vc G}_0$) is 
\begin{align}
P(\Delta{\vc G}_0|\vc G_0, \sigma,\{\vc x_i\})&\propto 
\exp\left(-\frac{\Delta{\vc G}_0^\tp\mat S\Delta{\vc G}_0}{2\sigma^2/N}\right)\delta({\vc G}_0^\tp\mat C\Delta{\vc G}_0)\nonumber\\
&\propto \exp\left(-\frac{N}{2\sigma^2}{\widehat{\vc G}_0^\tp\mat S\widehat{\vc G}_0}\right)\delta(\widehat{\vc G}_0^\tp\mat C\widehat{\vc G}_0-1)\ ,
\label{eq:sampling}\end{align}
where the delta-function enforces normalisation, 
ensuring that the resulting covariance matrix is 
proportional to the pseudoinverse of $\mat S$, not the (divergent) full inverse.  
The individual point errors $\Delta\vc x_i$ are not present in \eqref{eq:sampling}
other than via their contributions to the scatter matrix,
which is a sufficient statistic for this problem. 
The curvature bias correction has not been explicitly represented:
$\widehat{\vc G}_0$ here is the \emph{uncorrected} vector of coefficients.

The results to this point are in practice very similar to those of \cite{kanatani2015hyper},
although there are two significant differences in approach.
The first difference is that in this paper I have minimised the bias from each source separately:
the normalisation matrix is chosen to remove normalisation bias; 
reweighting bias is minimised by evaluating the gradient 
at a point consistent with the current best-fit model; 
and the curvature bias is corrected as a separate, final step. 
In contrast, the method of \cite{kanatani2015hyper} chooses the normalisation matrix 
to minimise all three biases simultaneously, including that from Sampson reweighting.  
One advantage of separate treatment is that 
the normalisation matrix may be calculated by the same method both before and after reweighting. 
The second difference is that I have taken my perturbation expansion 
only to the first nonvanishing order, 
which in some cases is first order and in others second.
The consistently second-order treatment of \cite{kanatani2011hyper} 
finds an additional normalisation bias term, 
but this is smaller than the leading term by a factor of $N$ 
and hence in the perturbative regime is always smaller than the the random error 
and not statistically significant.

\subsection{Posterior probabilities}
Although sampling distributions are useful for the general comparison of fitting methods, 
what we really want for an individual set of real measurements is 
not the distribution \eqref{eq:sampling}, 
which presumes knowledge of the true values that we do not have, 
but the posterior probability $P(\vc G|\{\widehat{\vc x}_i\})$.
For the Gaussian case the posterior probability and the sampling distribution are 
to first order interchangeable for a sufficiently broad prior, 
leading us to estimate the former as
\begin{equation}
P(\vc G|\{\widehat{\vc x}_i\}\sigma) \sim 
\exp\left(-\frac{N}{2\sigma^2}{\vc G}^\tp\widehat{\mat S}{\vc G}\right)\delta(\vc G^\tp\widehat{\mat C}\vc G-1)\ .
\label{eq:optimalPosterior}\end{equation}

In fact, noting that the optimally weighted error statistic is 
equivalent to the geometric mean-square error, 
we can obtain \eqref{eq:optimalPosterior} directly from Bayes' theorem.
With a Gaussian prior for the measurement noise we have
\begin{align}
P(\widehat{\vc x}|\vc G_0\sigma) &= 
\int\!\! \d {\vc x}\, P(\widehat{\vc x}|\vc x\sigma)P(\vc x|\vc G_0)\nonumber\\
&= \frac1{(\sqrt{2\pi}\sigma)^\Lambda}\int\!\!\d {\vc x}\, 
\exp\left(-\frac1{2\sigma^2}|\widehat{\vc x}-\vc x|^2\right)
P(\vc x|\vc G_0)\ .\\
\label{eq:measurementNoise}\end{align}
The prior probability for the (unknown) true point $\vc x$ may be factored into 
a delta-function ensuring the point is on the surface $Z(\vc x)=0$ and 
the distribution $\widetilde P$ of observed points over the surface,
\begin{equation}
P(\vc x|\vc G_0)= |\vcs\nabla Z(\vc x)|\delta(Z(\vc x))\widetilde P(\vc x|\vc G_0)\ .
\end{equation}
Now write the true point in terms of the point $\overline{\vc x}$ 
nearest to the measured $\widehat{\vc x}$ 
that satisfies the model and an offset $\vcs\xi$,
\begin{equation}
\vc x = \overline{\vc x} + \vcs\xi\ ;
\end{equation}
for $\vc x$ to be on the surface, we must have (to second order in $\vcs\xi$)
\begin{equation}
Z(\vc x) \simeq Z(\overline{\vc x}) + \vcs\xi\vcs.\vcs\nabla Z(\overline{\vc x})
+ \tfrac12 \left(\vcs\xi\vcs.\vcs\nabla\right)^2 Z(\overline{\vc x}) = 0\ .
\end{equation}
Since $\widehat{\vc x}-\overline{\vc x}$ is parallel to $\vcs\nabla Z(\overline{\vc x})$,
we can expand
\begin{align}
|\widehat{\vc x}-\vc x|^2 &= |\widehat{\vc x} - \overline{\vc x} - \vcs\xi|^2 \nonumber\\
&= |\widehat{\vc x} - \overline{\vc x}|^2 - 2(\widehat{\vc x} - \overline{\vc x})\vcs.\vcs\xi + |\vcs\xi|^2 \nonumber\\
&= |\widehat{\vc x} - \overline{\vc x}|^2 - 2\frac{(\widehat{\vc x} - \overline{\vc x})\vcs.\vcs\nabla Z(\overline{\vc x})}{|\vcs\nabla Z(\overline{\vc x})|^2}\vcs\nabla Z(\overline{\vc x})\vcs.\vcs\xi + |\vcs\xi|^2 \nonumber\\
&= |\widehat{\vc x} - \overline{\vc x}|^2 + 
\frac{(\widehat{\vc x} - \overline{\vc x})\vcs.\vcs\nabla Z(\overline{\vc x})}{|\vcs\nabla Z(\overline{\vc x})|^2}
\left(\vcs\xi\vcs.\vcs\nabla\right)^2 Z(\overline{\vc x}) + |\vcs\xi|^2 \nonumber\\
& = |\widehat{\vc x} - \overline{\vc x}|^2 + \vcs\xi^\tp\mat B \vcs\xi\ ,
\end{align}
where the elements of the matrix $\mat B$ are
\begin{equation}
B_{\alpha\beta} = \delta_{\alpha\beta} 
+ \frac{(\widehat{\vc x} - \overline{\vc x})\vcs.\vcs\nabla Z(\overline{\vc x})}{|\vcs\nabla Z(\overline{\vc x})|^2}
\partial_\alpha\partial_\beta Z(\overline{\vc x})\ .
\end{equation}
Substituting back into \eqref{eq:measurementNoise}, and 
taking $\widetilde P$ to be slowly varying on the typical scale of $\Delta\vc x$,
we have
\begin{align}
P(\widehat{\vc x}|\vc G_0\sigma) &\simeq 
 \frac1{(\sqrt{2\pi}\sigma)^\Lambda} 
\exp\left(-\frac1{2\sigma^2}|\widehat{\vc x}-\overline{\vc x}|^2\right)
\widetilde P(\overline{\vc x}|\vc G_0) \nonumber\\
&\quad \times \int\!\!\d {\vcs\xi}\, \exp\left(-\tfrac12\vcs\xi^\tp\mat B \vcs\xi\right)|\vcs\nabla Z(\overline{\vc x} + \vcs\xi)|\delta(Z(\overline{\vc x} + \vcs\xi)) \nonumber\\
&= \frac1{\sqrt{2\pi\sigma^2|\mat B|} }
\exp\left(-\frac1{2\sigma^2}|\widehat{\vc x}-\overline{\vc x}|^2\right)
\widetilde P(\overline{\vc x}|\vc G_0)
\ .\label{eq:surfaceIntegral}
\end{align}
To leading order in $\widehat{\vc x} - \overline{\vc x}$,
\begin{align}
|\mat B| &\simeq 1 + \frac{(\widehat{\vc x} - \overline{\vc x})\vcs.\vcs\nabla Z(\overline{\vc x})}{|\vcs\nabla Z(\overline{\vc x})|^2}
\nabla^2 Z(\overline{\vc x}) \nonumber\\
&\simeq\exp\left(\frac{(\widehat{\vc x} - \overline{\vc x})\vcs.\vcs\nabla Z(\overline{\vc x})}{|\vcs\nabla Z(\overline{\vc x})|^2}\nabla^2 Z(\overline{\vc x})\right)\ ,
\end{align}
giving
\begin{align}
P(\widehat{\vc x}|\vc G_0\sigma) &\simeq 
 \frac1{\sqrt{2\pi}\sigma }
\exp\left(-\frac1{2\sigma^2}\left|\widehat{\vc x}-\overline{\vc x}\vphantom{\frac{\sigma^2}{2}}+\frac{\sigma^2}{2}\frac{\vcs\nabla Z(\overline{\vc x})}{|\vcs\nabla Z(\overline{\vc x})|^2}\nabla^2 Z(\overline{\vc x})\right|^2\right)
\widetilde P(\overline{\vc x}|\vc G_0)
\ .\label{eq:probDensity}
\end{align}

We see that the density of measured points is higher on the concave side of the surface $Z=0$, 
which is an intuitively reasonable result.  
It is not however directly useful. 
To use Bayes' theorem to interchange the roles of $\widehat{\vc x}$ and $\vc G_0$ 
we must work with probabilities, not probability densities 
(as Section 15.7 of \cite{jaynes2003probability} illustrates).
The relevant volume element is not $\d^\Lambda {\vc x}$ but 
$\d|\widehat{\vc x}-\overline{\vc x}| \d^{\Lambda-1} \overline{\vc x}$, or equivalently  
$\d Z(\widehat{\vc x}) \d^{\Lambda-1} \overline{\vc x}$, 
factorised into the measured error (either geometric or algebraic) 
and the position on the surface, which was already integrated out in \eqref{eq:surfaceIntegral}.
We find that
\begin{align}
\mathrlap{P(\widehat{Z}|\vc G_0\sigma) \d \widehat{Z}}\qquad\nonumber\\&= 
\frac 1{|\vcs\nabla \widehat{Z}|}P(\widehat{\vc x}|\vc G_0\sigma)\d \widehat{Z}\nonumber\\
&\simeq
 \frac{\d{\widehat{Z}}}{\sqrt{2\pi}\sigma |\vcs\nabla \overline Z|
 \left(1+\left((\widehat{\vc x}-\overline{\vc x})\vcs.\vcs\nabla\right)\vcs\nabla\overline Z\right)}
\exp\left(-\frac1{2\sigma^2}\frac{\left|\widehat{Z}
+\frac{\sigma^2}{2}\nabla^2 \overline Z\right|^2}{|\vcs\nabla \overline Z|^2}\right)\nonumber\\
&\simeq
 \frac{\d{\widehat{Z}}}{\sqrt{2\pi}\sigma |\vcs\nabla \overline Z| }
\exp\left(-\frac1{2\sigma^2}\frac{\left|\widehat{Z}
-\frac{\sigma^2}{2}\nabla^2 \overline Z\right|^2}{|\vcs\nabla \overline Z|^2}\right)
\ ,
\end{align}
where we have expanded the gradient and again used the fact that 
$\widehat{\vc x}-\overline{\vc x}$ is parallel to $\vcs\nabla Z(\overline{\vc x})$,
more specifically that 
\begin{align}
\widehat{\vc x}-\overline{\vc x} \simeq \frac{\vcs\nabla \overline Z}{|\vcs\nabla \overline Z|^2}\widehat{Z}\ .\label{eq:probZ}
\end{align}

\begin{figure}[tb]
\centering\includegraphics[width=1.00\textwidth]{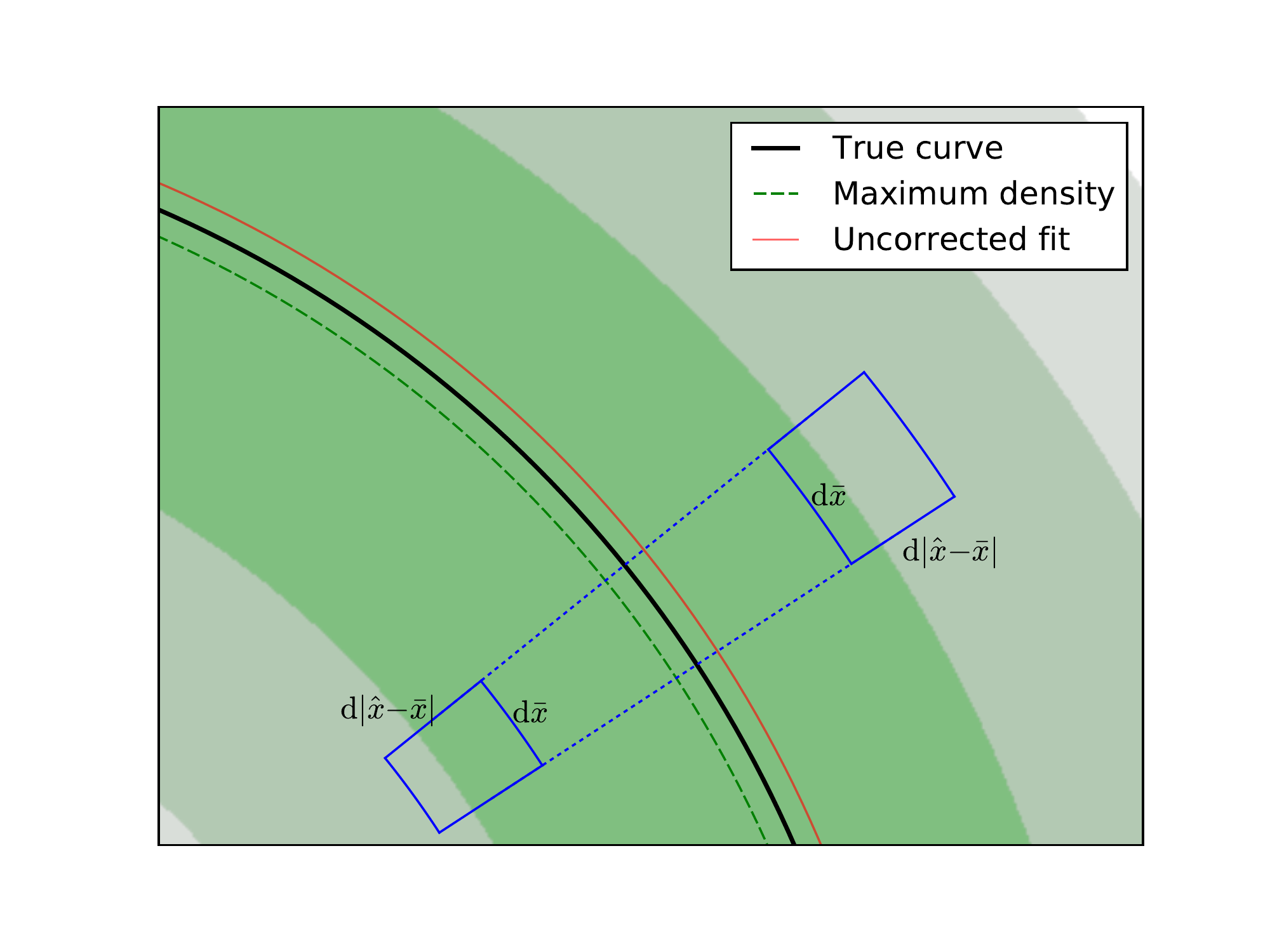}
\caption{\label{fig:curvature}
The probability density $P(\widehat{\vc x})$ (indicated by the shading) is larger on the concave side of the curve,
but for each element of the curve $\d \overline{x}$ 
the corresponding area $\d A = \d \overline{x} \d |\widehat{\vc x}-\overline{\vc x}|$
is larger on the convex side,
and so is the element of probability $P(\widehat{\vc x})\d A$. 
Thus the uncorrected fit lies outside the true curve.
}
\end{figure}

Thus for probabilities the result of \eqref{eq:probDensity} is reversed:
for any given small area of the surface, 
there is a greater probability of finding a nearby point on the convex side than on the concave.
This is illustrated for the 2D case in Figure \ref{fig:curvature}.  
Taking the product over all sample points and using Bayes' theorem 
to convert  $P(\{\widehat{Z}_i\}|\vc G_0\sigma)$ to  $P(\vc G|\{\widehat{Z}_i\}\sigma)$ 
now directly yields \eqref{eq:optimalPosterior} 
(including an explicit representation of the curvature correction).

While the perturbative results obtained earlier show that the method recommended in this paper is 
optimal among the class of linear algebraic methods,
the direct derivation demonstrates the stronger result that it is globally optimal:
within the small-noise approximation, no method gives a superior predictor of ${\vc G}$
(though other methods may be equally good).

Since we do not know in advance the value of $\sigma$, we integrate over 
the Jeffreys prior \cite{jeffreys1939theory,jaynes2003probability}
$P(\sigma)\propto1/\sigma$ to obtain a $t$-distribution,
\begin{equation}
P(\vc G|\{\widehat{\vc x}_i\}) \sim 
\left({\vc G}^\tp\widehat{\mat S}{\vc G}\right)^{-N/2}\delta(\vc G^\tp\widehat{\mat C}\vc G-1)\ ;
\end{equation}
for large $N$ this reduces back to 
\begin{equation}
P(\vc G|\{\widehat{\vc x}_i\}) \sim 
\exp\left(-\frac{N}{2\widehat\sigma^2}{\vc G}^\tp\widehat{\mat S}{\vc G}\right)\delta(\vc G^\tp\widehat{\mat C}\vc G-1)\ ,
\label{eq:measuredPosterior}\end{equation}
where $\widehat\sigma^2 = \widehat{\vc G}_0^\tp\widehat{\mat S}\widehat{\vc G}_0$.
The corresponding covariance matrix can be found from \eqref{eq:covariance} 
with the obvious substitutions of measured for true values.

These results hold for the case of a broad prior distribution for $\vc G$,
when the posterior distribution is just the normalised likelihood.
If we wish to impose additional constraints on the solution, 
we can introduce a suitable restrictive prior 
and multiply it by the likelihood \eqref{eq:measuredPosterior} 
to obtain the corresponding constrained posterior probability.

\section{Application to conic sections}

For conic fitting we have 2-dimensional data points $\vc x^\tp = \begin{pmatrix}x & y\end{pmatrix}$ (in Cartesian coordinates), with
the 6-component design vector 
\begin{equation}
\vc D^\tp(\vc x) = \begin{pmatrix}x^2&x y& y^2&x& y&1\end{pmatrix}\ .
\end{equation}
The self-normalising constraint matrix $\mat C_\text{N}$ is of rank 5, and is already in the desired form \eqref{eq:partitionC};
it can be constructed directly from the elements of $\mat S$
using eqn(75) of  \cite{collett2014ellipse}.
The constant rank-3 constraint matrices preferred in earlier treatments \cite{bookstein1979fitting, fitzgibbon1999direct} 
inevitably result in biased estimates, and will not be considered further in this paper; 
the biases are exhibited numerically in \cite{collett2014ellipse} 
(note that the results therein include some 
for highly eccentric ellipses that are beyond 
the regime in which the perturbation theory used in this paper is accurate).  

The curvature bias can indeed be expressed in the form \eqref{eq:linearLaplace},
with $\mat L$ having nonzero elements only in its last row,
which is $\begin{pmatrix}1 & 0 & 1 & 0 & 0 & 0\end{pmatrix}$.
From this it follows that $\mat L \mat C=\mat C\mat L^\tp=\mat 0$, 
so the curvature bias correction does not alter the normalisation
and the residual curvature bias \eqref{eq:residCurv} vanishes completely.
Furthermore, the approximate equality in \eqref{eq:curvCorrect} is exact, 
so actually performing the bias correction is simply a matter of 
modifying the final component $g_{n6}$ of each eigenvector $\vc G_n$ to
\begin{equation}
g_{n6}' = g_{n6} + \widehat\sigma^2(g_{n1}+g_{n3})\ ,
\label{eq:conicCorrection}\end{equation}
where $\widehat\sigma^2 = \lambda_0$.

For iterative reweighting according to \eqref{eq:optimalWeight} 
we may evaluate the gradient $\vc D_{i,\mu}$ at a point on the estimated curve 
determined with the aid of the appropriate elliptical coordinate system \cite{collett2014ellipse} 
(or confocal parabolic coordinates if the curve is a parabola, 
but this is vanishingly likely to happen for real data unless we force it).
Introduce the coordinates $(\eta, \theta)$ such that
\begin{equation}
\vc x = \cos\theta\cosh\eta\,\vc f_\parallel + \sin\theta\sinh\eta\,\vc f_\perp + \vc c\ ,
\label{eq:ellipticalCoords}\end{equation}
where the focal points are at $\vc c \pm \vc f_\parallel$, with $\vc f_\perp = \begin{pmatrix}-f_{\parallel y} & f_{\parallel x}\end{pmatrix}^\tp$;
the current estimated fit is a curve of constant $\eta$ if it is an ellipse 
or of constant $\theta$ if it is a hyperbola.  
Each measured point $\widehat{\vc x}_i$ can be expressed in these coordinates as  
$(\widehat\eta_i, \widehat\theta_i)$; 
the corresponding best estimate $\overline{\vc x}_i$ is 
at $(\eta, \widehat\theta_i)$ for the elliptical case 
or $(\widehat\eta_i, \theta)$ for the hyperbolic.   

\subsection{Numerical comparisons}

\begin{figure}[tb]
\centering\includegraphics[width=1.00\textwidth]{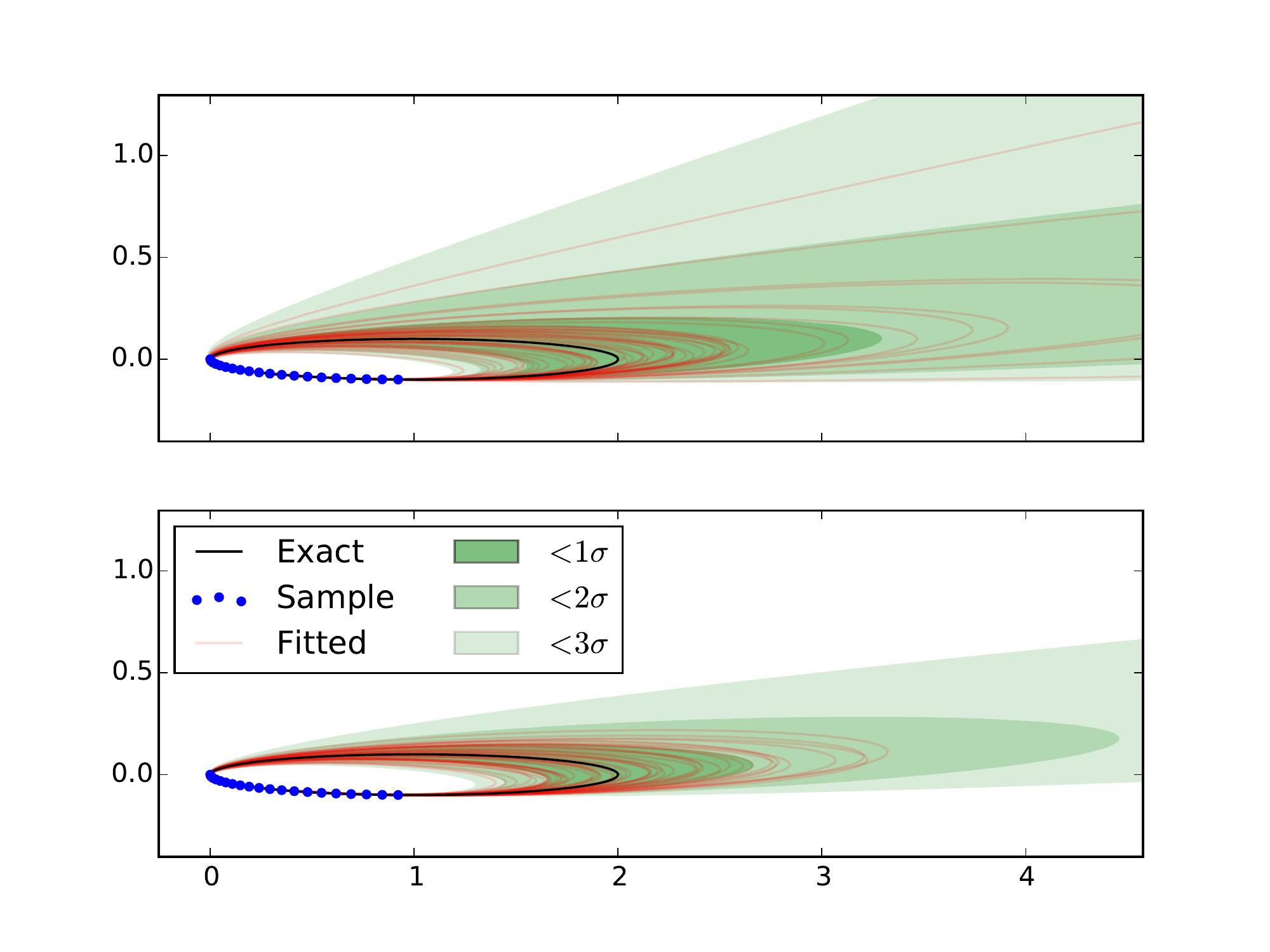}
\caption{\label{fig:generic} Comparison of predicted and observed variability in fitting.
An ellipse with semimajor axis $1.0$ and semiminor axis $0.1$ is 
sampled 50 times, each sample containing 20 points distributed along one quadrant 
with individual measurement error $0.001$.
The upper subfigure is for unweighted data and the lower for weighted.}
\end{figure}

\begin{figure}[tb]
\centering\includegraphics[width=1.00\textwidth]{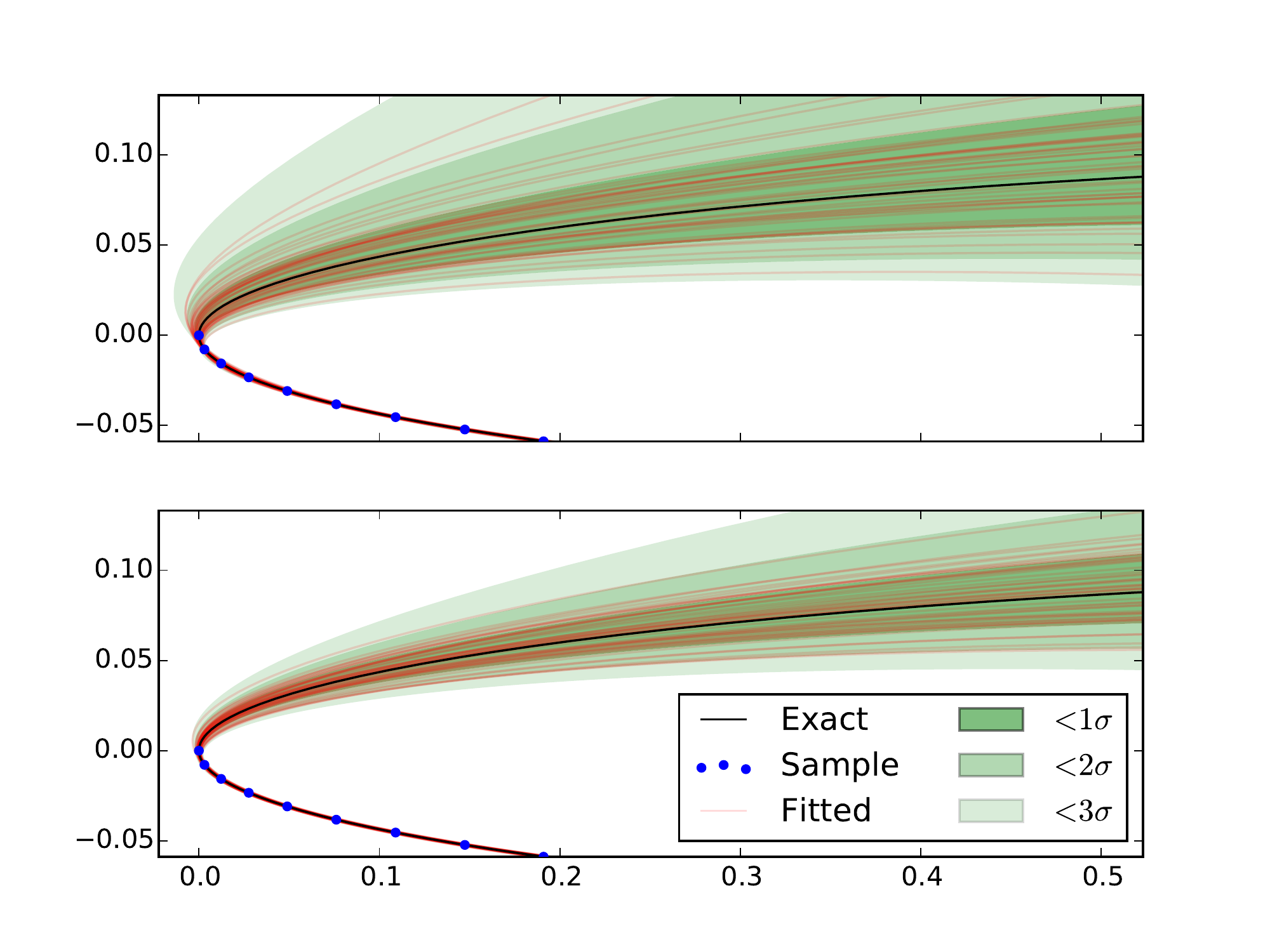}
\caption{\label{fig:generic_closeup} A close-up of Figure \ref{fig:generic}, 
showing the region in which the variability of the fitted curve increases 
as it moves away from the data points.}
\end{figure}

Figures \ref{fig:generic} and \ref{fig:generic_closeup} compare 
the sampling distribution predicted by \eqref{eq:covariance} with
a number of individual fits to randomly generated data.
The original curve is an ellipse with semimajor axis $1.0$ and semiminor axis $0.1$, 
sampled at 20 points distributed along one quadrant, 
each sample point having a random error of standard deviation $0.001$ in each direction.
The fitting uses the self-normalising method (i.e.\ $\mat C = \mat C_\text{N}$)
with the simple curvature correction \eqref{eq:conicCorrection}. 
The shaded confidence intervals are bordered by contours of
\begin{equation}
\frac{Z(\vc x)}{\sqrt{\operatorname{Var}(Z(\vc x))}}
=\frac{\vc G^\tp \vc D(\vc x)}{\sqrt{\vc D^\tp(\vc x) \mat V_0 \vc D(\vc x)}}\ , 
\end{equation}
where $\vc G$ is the vector of coefficients corresponding to the exact curve.
In the upper subfigure both the fits and the intervals are calculated using unweighted 
(i.e.\ equally weighted) data; 
in the lower subfigure, the fits are iterated with reweighting,
and the intervals are calculated from the ideal weighting, using \eqref{eq:optimalCovariance}.
There are 50 individual sample sets, so we expect that at any given point around the ellipse, 
typically 2 or 3 will lie outside the $2\sigma$ confidence interval,
and about 14 between $1\sigma$ and $2\sigma$;
the observed results are  consistent with this expectation, 
for both unweighted and weighted fits.
We see that both predicted and observed variability is 
reduced by about a factor of 2 by the reweighting;
it should be stressed that this does not mean that 
any individual fit is guaranteed to be improved by reweighting 
(it is perfectly possible to find examples that are made significantly worse), 
only that the weighted fit is more precise on average.
For these parameters, the unweighted fit changes from elliptical to hyperbolic 
somewhere between 2 and 3 standard deviations outside the exact result,
while the weighted fit is still elliptical at $3\sigma$.

\begin{figure}[tb]
\centering\includegraphics[width=1.00\textwidth]{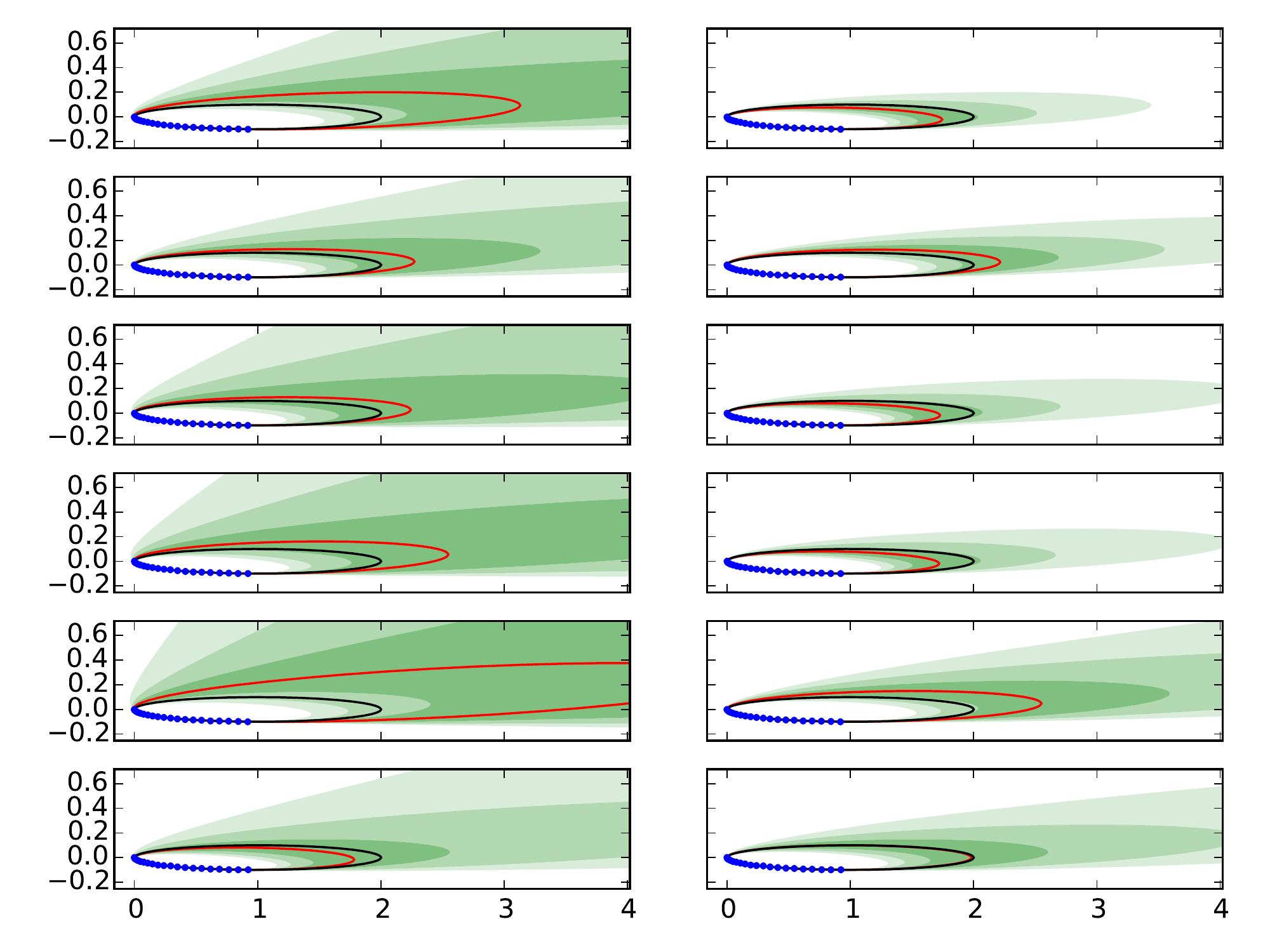}
\caption{\label{fig:posterior} 
Posterior confidence intervals for a few of the samples from Figure \ref{fig:generic}.  
Unweighted fits are on the left, and corresponding weighted ones on the right.
(Shading key as in Fig.\ \ref{fig:generic}.)}
\end{figure}

Figure \ref{fig:posterior} shows the best fit and confidence intervals
estimated from a few individual samples using \eqref{eq:covariance}.
Again the results are consistent with expectations: 
for the majority of samples the true curve is within the $1\sigma$ confidence interval, 
and for the rest it is within the $2\sigma$ interval; 
most but not all individual fits are improved by iteration with reweighting.

\begin{figure}[tb]
\centering\includegraphics[width=1.00\textwidth]{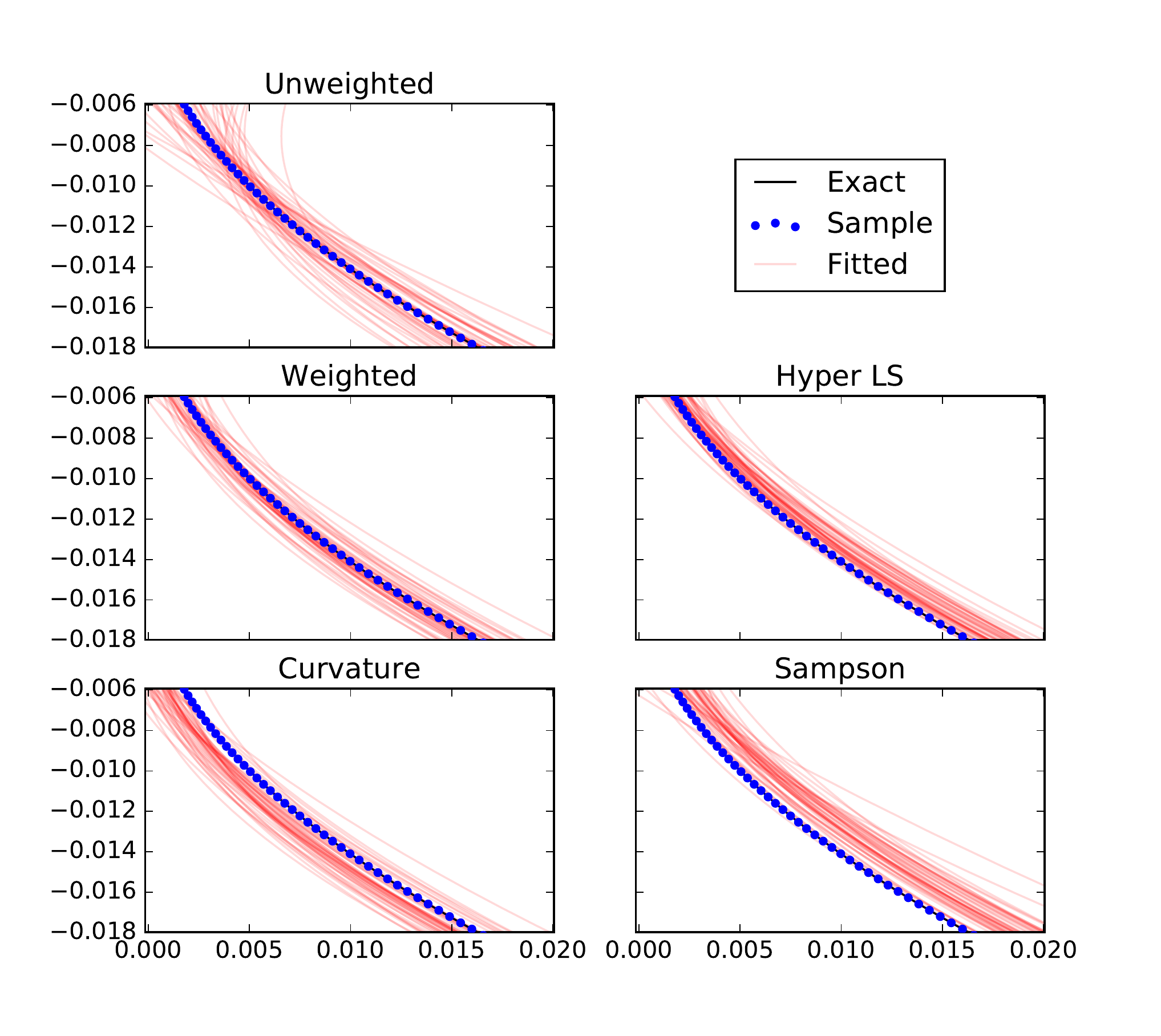}
\caption{\label{fig:bias} 
An ellipse with semimajor axis $1.0$ and semiminor axis $0.1$ is 
sampled 50 times, each sample containing 500 points distributed along one quadrant 
with individual measurement error $0.004$.
The top subfigure shows an unweighted fit (with curvature correction), 
the centre-left an unbiased weighted fit, 
the lower left a weighted fit without curvature bias correction
and the lower right one weighted by Sampson's gradient method.
The centre-right shows the result of a `hyper-renormalised' fit 
following the method of \cite{kanatani2015hyper}.
}
\end{figure}

For the parameters used for Figs \ref{fig:generic}--\ref{fig:posterior}, 
the normalisation bias from a fixed normalisation leads to very obvious fitting errors,
as illustrated in \cite{collett2014ellipse}.
The effects of the curvature bias and the bias from Sampson's reweighting are much smaller,
but can be seen by increasing the number of sample points and the individual measurement error
(still within the perturbative regime), as shown in Fig.\ \ref{fig:bias}.
  
Omitting the curvature bias correction results in 
fits falling outside the true curve near the tip (where the curvature is largest);
reweighting using the gradient at the measured points instead of on the estimated curve
results in fits that lie inside the true curve over a rather larger region.
A geometric fit using a computationally expensive nonlinear least-squares method would produce 
results very similar to those obtained by omitting the curvature bias correction.
The `hyper-renormalised' method of \cite{kanatani2015hyper} uses the Sampson weighting 
but modifies the normalisation matrix to correct for this: 
as Fig.\ \ref{fig:bias} shows, this reduces the weighting bias but does not completely eliminate it.

\section{Type-specific fitting}
\subsection{Parabolic constraint}
\label{sec:parabolic}

To enforce a parabolic solution, we may use the likelihood as obtained previously
in \eqref{eq:optimalPosterior},
but combine it with a prior probability that respects the type-specific constraint,
giving the posterior probability with optimal weighting as
\begin{multline}
P(\vc G|\{\widehat{\vc x}_i\}) \\\sim 
\exp\left(-\frac{N}{2\widehat\sigma^2}{\vc G}^\tp\widehat{\mat S}{\vc G}\right)\delta(\vc G^\tp\widehat{\mat C}\vc G-1)\delta(\vc G^\tp\mat Q\vc G)\ ,\\
\label{eq:parabolicPosterior}\end{multline}
where 
\begin{equation}
{\vc G}^\tp\mat Q{\vc G} = 4 g_1 g_3- g_2^2\ .
\end{equation}

Clearly, the best estimate (both the most probable value and the posterior mean) 
is the value $\overline{\vc G}$ satisfying the constraint that is 
the shortest Mahalanobis distance from the unconstrained solution.
That is, we seek the coefficient vector $\overline{\vc G}$ that minimises
$\overline{\vc G}^\tp\widehat{\mat S}\overline{\vc G}$ subject to the constraint that
$\overline{\vc G}^\tp\mat Q\overline{\vc G} = 0$,
while also preserving the normalisation constraint that $\overline{\vc G}^\tp\widehat{\mat C}\overline{\vc G}=1$.
As a first-order approximation to $\overline{\vc G}$, we have
\begin{equation}
\overline{\vc G} \simeq \vc G_0 
- \frac{{\vc G_0}^\tp\mat Q{\vc G_0}}{2\vc G_0^\tp\mat Q\mat Y_0\mat Q\vc G_0}\mat Y_0\mat Q\vc G_0\ ,
\label{eq:approxParabola}\end{equation}
where $\vc G_0$ is the unconstrained solution.  
This may be iteratively refined by 
alternately projecting $\overline{\vc G}-\vc G_0$ 
onto the normal to the surface of constant $\overline{\vc G}^\tp\mat Q\overline{\vc G}$ 
(to ensure that the length is minimised) and reapplying \eqref{eq:approxParabola} 
with the current best estimate of $\overline{\vc G}$ replacing $\vc G_0$, repeating until
$\overline{\vc G}^\tp\mat Q\overline{\vc G}$ is sufficiently close to zero.

Linearising the type-specific constraint around this best point estimate gives a residual Gaussian distribution of the form
\begin{equation}
P(\Delta\vc G|\{\widehat{\vc x}_i\}) \sim 
\exp\left(-\frac{N}{2\widehat\sigma^2}{\Delta\vc G}^\tp \overline{\mat S}{\Delta\vc G}\right)\ ,
\label{eq:parabolicResidual}\end{equation}
where
\begin{align}
\overline{\mat S} &= (1-\mat P_0^\tp-\overline{\mat P}{}^\tp)\widehat{\mat S}(1-\mat P_0- \overline{\mat P})\ ;\label{eq:parabolicS}
\end{align}
the projectors
\begin{equation}
\mat P_0 = \vc G_0\vc G_0^\tp\widehat{\mat C}\quad\text{and}\quad
\overline{\mat P} = \frac{\widehat{\mat C}{\vphantom{(}}^{+}\mat Q\overline{\vc G}\,\overline{\vc G}^\tp\mat Q}{\overline{\vc G}^\tp\mat Q\widehat{\mat C}{\vphantom{(}}^{+}\mat Q\overline{\vc G}}
\end{equation}
ensure that only deviations respecting both the normalisation and parabolic constraints respectively are considered. 
The covariance matrix is proportional to 
the generalised inverse $\overline{\mat Y}$ of the rank 4 matrix $\overline{\mat S}$,
calculated in the fashion specified by \eqref{eq:pseudoinverse}. 
We see that obtaining accurate confidence intervals 
is numerically more expensive for the parabolic fit than for the generic one.

\begin{figure}[tb]
\centering\includegraphics[width=1.00\textwidth]{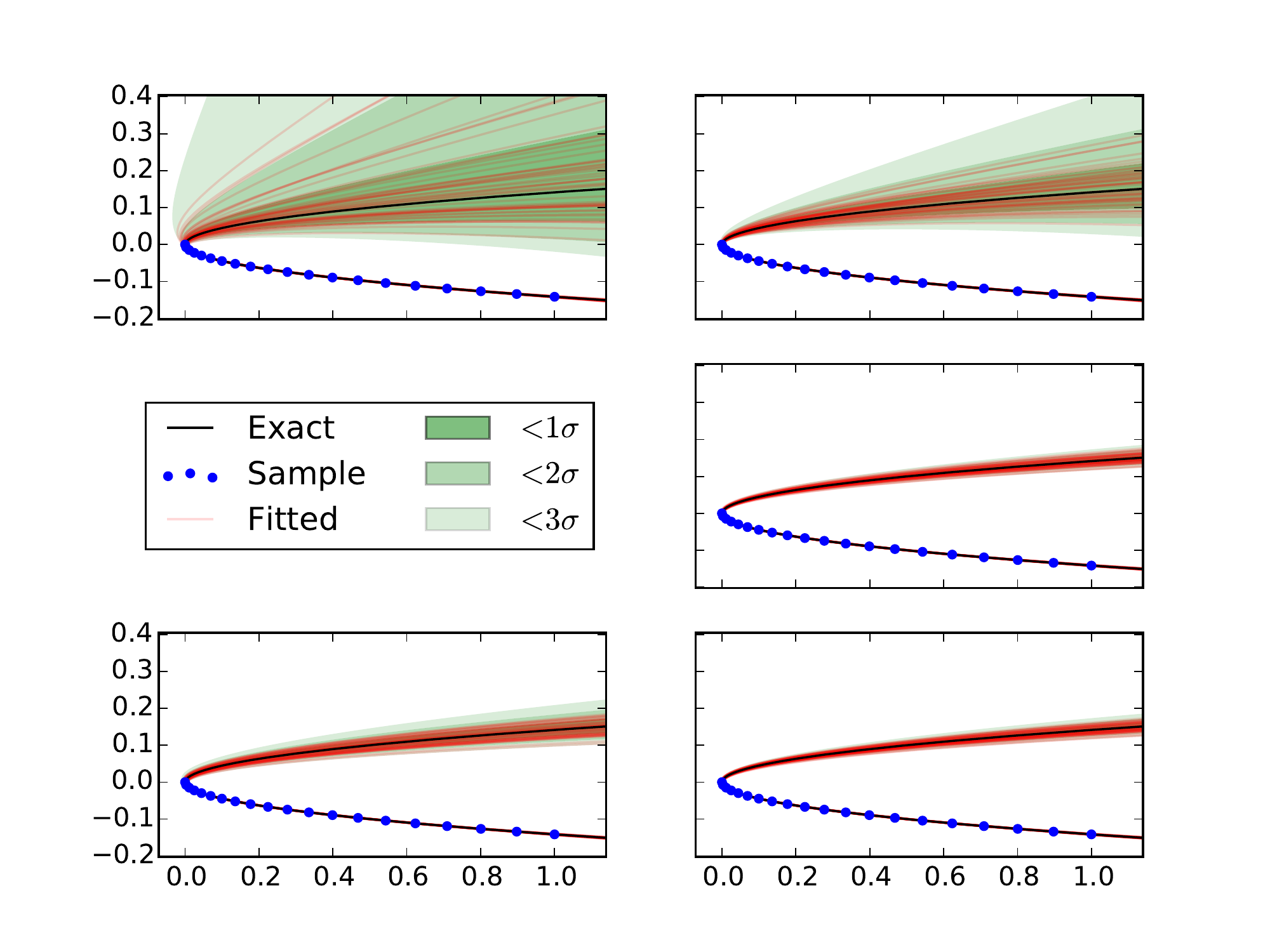}
\caption{\label{fig:parabolic} Comparison of variability in generic and type-specific parabolic fitting.
A parabola was sampled and fitted 50 times, each sample containing 20 points 
with individual measurement error $0.001$.
The top plots are unconstrained generic fits, and the bottom ones constrained to be parabolic;
the left-hand ones are unweighted and the right-hand ones optimally weighted.
The centre-right plot is a hybrid case: a generic initial fit determined the weights but the weighted fit was type-specific.}
\end{figure}

Figure \ref{fig:parabolic} again shows both 
predicted sampling distributions and
a sample of individual fits to randomly generated data.
The original curve is a parabola with focal length $0.01$, 
sampled at 20 points distributed along one arm, 
each sample point having a random error of standard deviation $0.001$ in each direction.
Comparing the lefthand plots to the righthand ones, 
we see as previously a moderate reduction in fitting error from reweighting.
A greater improvement is apparent when comparing 
the type-specific fits in the bottom subfigures with 
the corresponding generic ones in the top subfigures.  
In all cases there is adequate agreement between predicted and observed variability.

The `hybrid' method (generic preliminary fit and type-specific final fit) used for the centre-right subfigure gives results that are 
indistinguishable from the fully type-specific method at the bottom right; 
that is, it does not matter whether 
the preliminary fit used for the weighting was type-specific or not.  
This is not surprising: 
even though the initial generic and parabolic fits are very different globally, 
they are close together in the region of the data points, so the resulting weights are similar; 
and only large changes in weighting have a detectable effect on the distribution of fitted curves.

\subsection{Elliptical or hyperbolic constraint}

The parabolic constraint is the equality ${\vc G}^\tp\mat Q{\vc G}=0$, 
giving a delta-function factor in \eqref{eq:parabolicPosterior}.  
To enforce an elliptic or hyperbolic solution requires instead the satisfaction of the corresponding \emph{in}equality,
leading to
\begin{align}
P(\vc G|\{\widehat{\vc x}_i\})\sim 
\exp\left(-\frac{N}{2\widehat\sigma^2}{\vc G}^\tp\widehat{\mat S}{\vc G}\right)\delta(\vc G^\tp\widehat{\mat C}\vc G-1)u(\pm\vc G^\tp\mat Q\vc G)\ ,
\label{eq:ellipticPosterior}\end{align}
where $u$ is the Heaviside unit step, with the sign factor for its argument positive for the elliptic case and negative for the hyperbolic. 
The residual distribution is no longer locally Gaussian as \eqref{eq:parabolicResidual} is,  
but is truncated along the direction normal to the surface ${\vc G}^\tp\mat Q{\vc G}=0$.

Thus the best estimate of $\vc G$, 
given by the posterior mean, is \emph{not} the most probable value,
but lies on a conic pencil passing through 
the unconstrained solution and the nearest point on the surface
(i.e.\ the best parabolic fit as calculated in the previous section).
Perpendicular to the pencil the distribution is still locally Gaussian, 
but along it we must account for the truncation.
Quantitatively, we have for a one-dimensional truncated Gaussian that
\begin{align}
\expect{x} &= \int_{x_0}^{\infty} x e^{-{x^2}/{2}}\d x \bigg/ \int_{x_0}^{\infty} e^{-{x^2}/{2}}\d x\nonumber\\
&= \sqrt{\frac2\pi}\frac{e^{-x_0^2/2}}{\operatorname{erfc}(x_0/\sqrt2)}\ .
\label{eq:truncatedGaussian}\end{align}
The estimated posterior mean for the type-constrained conic is accordingly
\begin{align}
\expect{\vc G} = \vc G_0 + \sqrt{\frac2\pi}\frac{e^{-x_0^2/2}}{\operatorname{erfc}(x_0/\sqrt2)}\frac{\overline{\vc G}-\vc G_0}{x_0}\ ,
\label{eq:specificMean}\end{align}
where $\vc G_0$ is the unconstrained estimate, $\overline{\vc G}$ is the nearest parabola,
and $|x_0|$ is the Mahalanobis distance between the two,
\begin{align}
x_0 = \pm\sqrt{N\left(\frac{\overline{\vc G}^\tp\widehat{\mat S}\,\overline{\vc G}}{\widehat\sigma^2}-1\right)}\ .
\label{eq:meanOffset}\end{align}

\begin{figure}[tb]
\centering\includegraphics[width=1.00\textwidth]{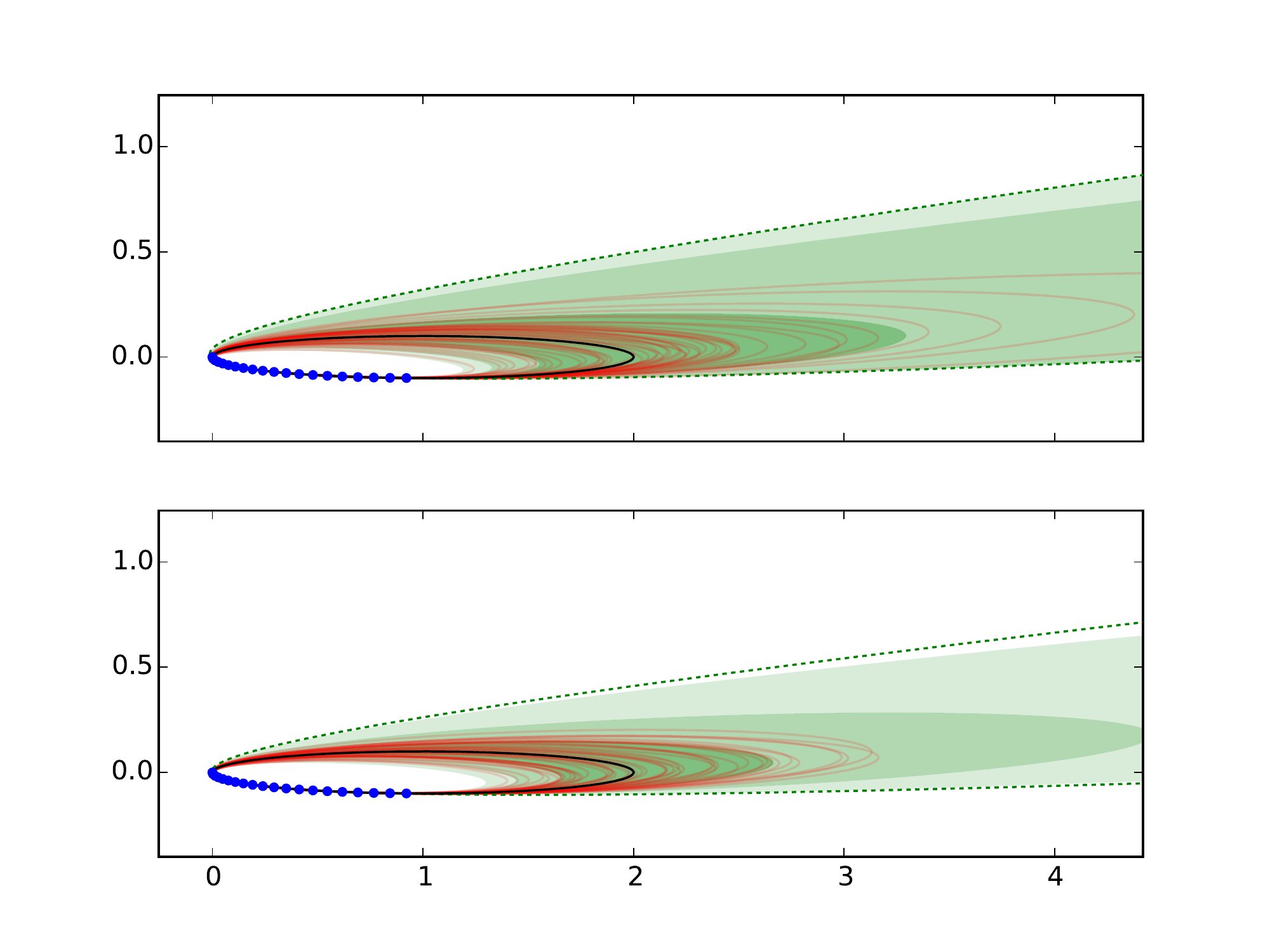}
\caption{\label{fig:elliptic}
Type-specific ellipse fitting, using the same samples as Fig.\ref{fig:generic}.
As previously, the upper subfigure is for unweighted data and the lower for weighted.
The dotted line is the parabola that best fits the noise-free sample points.
}
\end{figure}

\begin{figure}[tb]
\centering\includegraphics[width=1.00\textwidth]{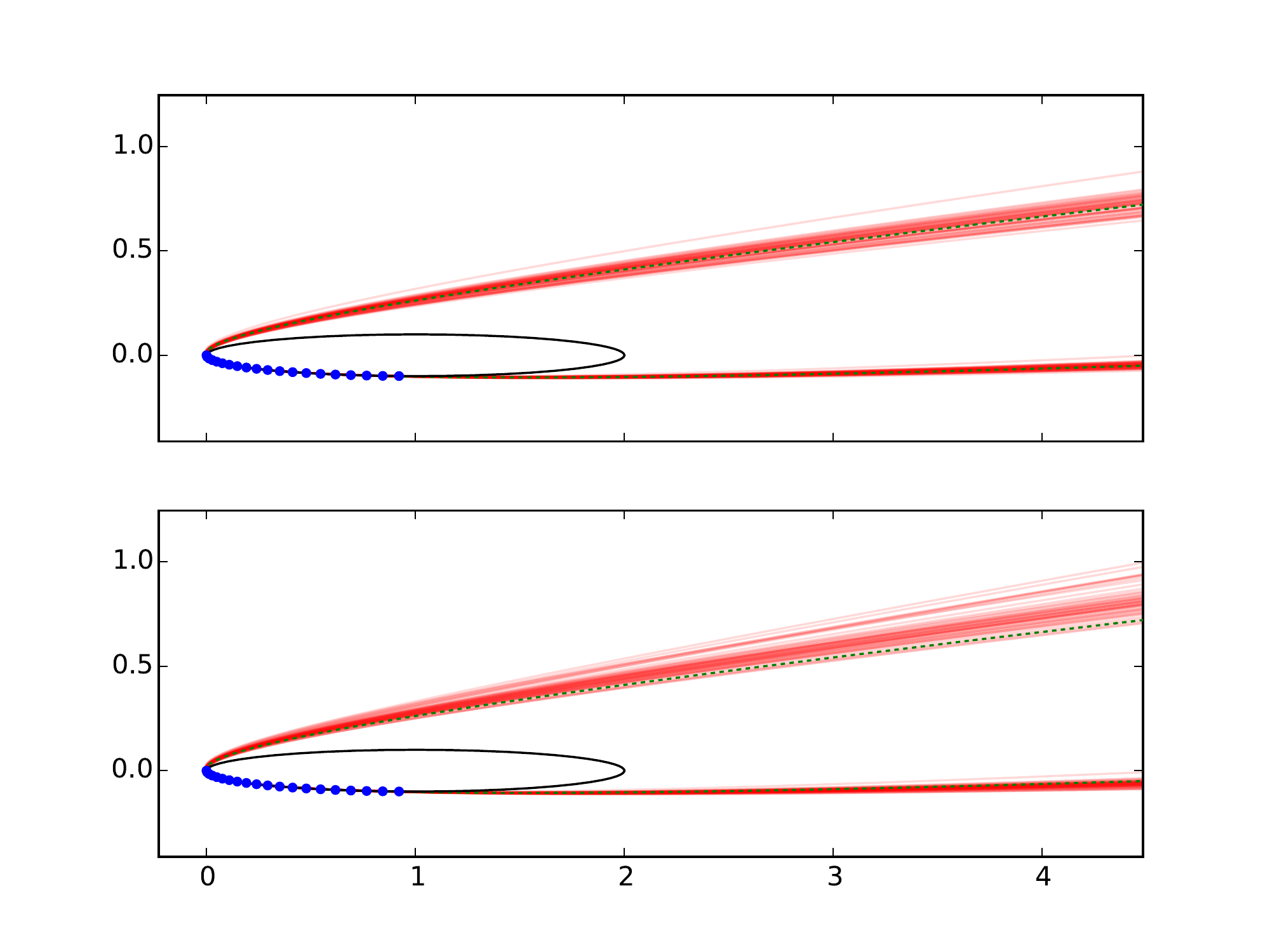}
\caption{\label{fig:hyperbolic}
Forcing elliptical data to fit a parabola (upper subfigure) or a hyperbola (lower subfigure).
The dotted line is again the parabola that best fits the noise-free sample points.
}
\end{figure}

There are two distinct cases.  
Usually, when the unconstrained solution is already of the correct type, 
we take the negative sign in \eqref{eq:meanOffset},
and the mean will be displaced a small distance along the pencil away from the boundary.
The first-order approximation \eqref{eq:approxParabola} for the boundary point is sufficiently accurate.
But if the unconstrained solution is of the wrong type, 
as may sometimes occur if the noise is large or the true solution is very close to the boundary, 
we need the positive sign in \eqref{eq:meanOffset},
and the mean will be at a point on the pencil close to the boundary, 
on the far side from the unconstrained solution.
In this case we may require iterative refinement of the boundary point estimate,
as for the parabolic constraint itself.
In neither case is the mean \emph{between} the two points defining the pencil.

It would in principle be possible to calculate 
the covariance matrix of an elliptical or hyperbolic solution
as we have done for the generic and parabolic cases. 
But in this case, because the distribution is not symmetric, 
it is both simpler and more informative to work directly with 
the full parameter set of the probability distribution,
namely the generic solution $\vc G_0$, the generic covariance matrix $\mat V_0$, 
and the parabolic solution $\overline{\vc G}$.
The first two give the same Gaussian as the generic case, 
and the last the surface along which that Gaussian is truncated.

Figure \ref{fig:elliptic} shows the results of ellipse-specific fits 
of the same samples as in Fig.\ref{fig:generic}.  
In the unweighted case, the truncation of the sampling distribution at the parabolic solution 
is significant, and the resulting fits have 
a noticeably smaller spread than the generic ones in Fig.\ref{fig:generic}.  
In the weighted case, for these parameters, the truncation occurs in the tail of the distribution, 
and the type-specific fits are accordingly little changed from the generic ones. 
This is appropriate behaviour for an unbiased type-specific method: 
if the generic fit is already clearly of the correct type, 
the type-specific requirement is redundant information 
and should not significantly alter the result.
(For simplicity of presentation, 
the figure shows the distribution in $\vc x$-space truncated along the path of the parabola, 
but this is only an approximation to the actual truncation of the distribution in $\vc G$-space.)

Figure \ref{fig:hyperbolic} shows what happens if the same elliptical data are 
erroneously forced to fit a parabola or hyperbola.  
As is to be expected, the parabolic solutions cluster closely around, 
and the hyperbolic ones closely outside of, the noise-free best-fit parabola.

\section{Other parameterisations}

So far the discussion of fitting errors has focused on the conic coefficients $\vc G$;
the figures also use contours of the algebraic error $Z$, which is a linear function of $\vc G$.
But there are other properties of the fitted curves that may be of interest, 
such as the location of the focal points, or the ratio of the axis lengths. 
The statistics of these may be estimated by standard error propagation techniques.

Consider a parameter $\rho$ given by some function $r$ of the components of $\vc G$.  In the presence of noise, we obtain a estimated value
\begin{align}
\widehat \rho &= r(\widehat{\vc G}) \nonumber\\
&\simeq r(\vc G) + \sum_m \frac{\partial r}{\partial g_m}\Delta g_m + \frac12\sum_{mn} \frac{\partial^2 r}{\partial g_m\partial g_n}\Delta g_m\Delta g_n \nonumber\\
&=\rho + \Delta\vc G^\tp\vc r' + \tfrac12\Delta\vc G^\tp \mat R''\Delta\vc G
\ ,
\end{align}
where $\vc r'$ and $\mat R''$ are the gradient and Hessian of $r$.  
Since we have ensured that $\Delta\vc G$ has zero mean, 
any bias in our estimate of $\rho$ comes from the mean of the second-order term, 
\begin{align}
\expect{\widehat \rho - \rho} = \tfrac12\tr(\mat R''\mat V)\ , 
\end{align}
while to leading order the variance of the estimate is
\begin{align}
\expect{\widehat \rho\,^2}- \expect{\widehat \rho\,}^2 = \vc r'^\tp\mat V\vc r'\ . 
\end{align}

\begin{figure}[tb]
\centering\includegraphics[width=1.00\textwidth]{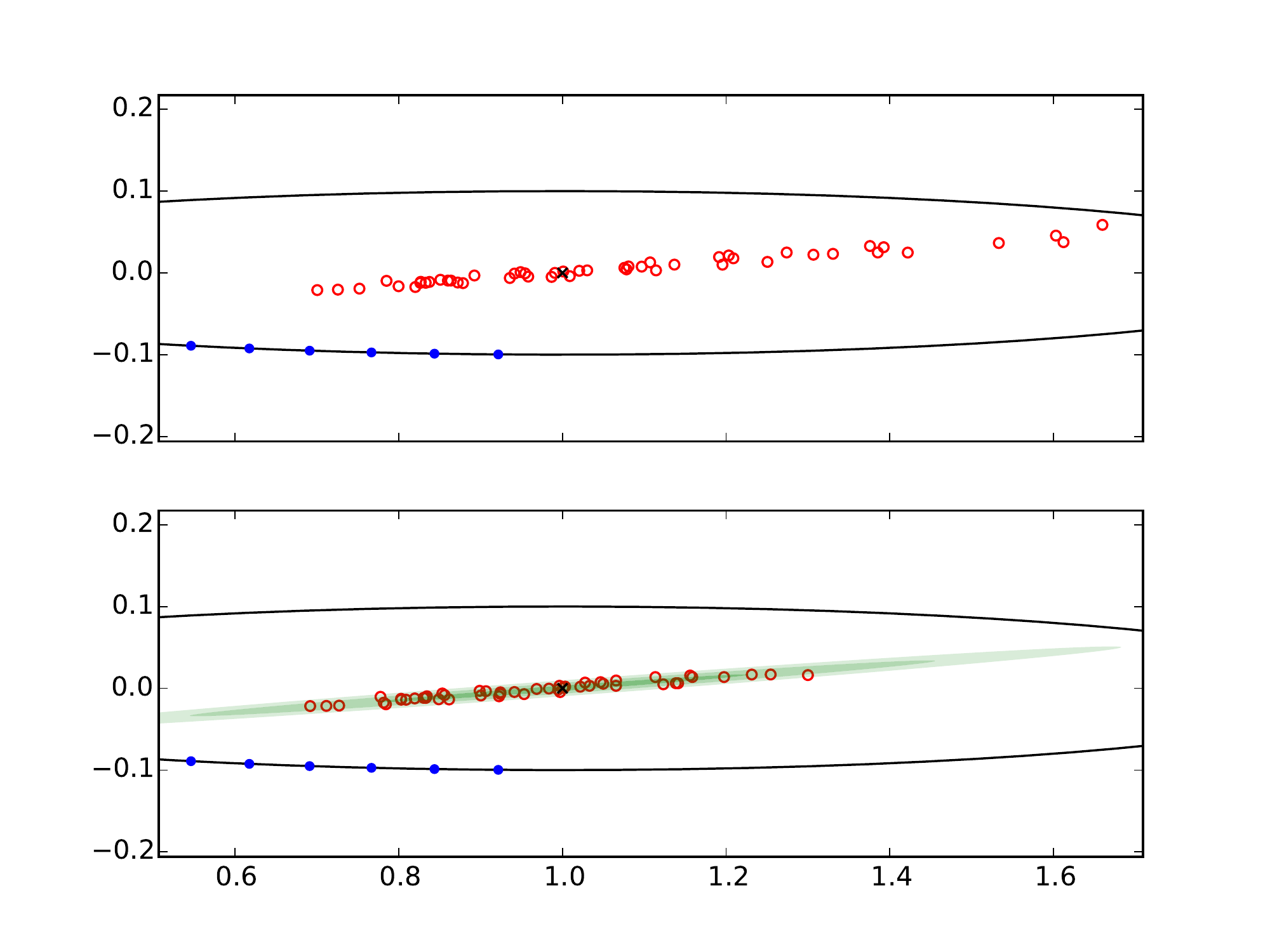}
\caption{\label{fig:centres}
Top: centres of fitted ellipses, as directly calculated.
Bottom: centres with bias correction, compared to the predicted sampling distribution.
Sample values are open circles, and `\textsf{x}' marks the actual ellipse centre;
the predicted sampling distribution is shaded as in earlier figures.
}
\end{figure}

For example, we may wish to find the centre of a fitted ellipse.
In the absence of noise, the centre point is
\begin{align}
\vc c 
= -{\begin{pmatrix} 2g_1 & g_2 \\ g_2 & 2g_3 \end{pmatrix}}^{-1}
\begin{pmatrix} g_4 \\ g_5 \end{pmatrix}\ .
\end{align}
With noise, the bias in our estimate is
\begin{align}
\expect{\Delta\vc c} = \frac12\sum_{mn}V_{mn}\frac{\partial^2 \vc c}{\partial g_m\partial g_n}\ ,
\label{eq:centreBias}\end{align}
and the covariance
\begin{align}
\expect{\Delta\vc c\Delta\vc c^\tp}-\expect{\Delta\vc c}\expect{\Delta\vc c}^\tp
= \sum_{mn}V_{mn}\frac{\partial \vc c}{\partial g_m}\frac{\partial \vc c^\tp}{\partial g_n}\ .
\label{eq:centreVariance}\end{align}
Figure \ref{fig:centres} illustrates the results for the same data set as used in previous figures.  
Without correcting for the bias, 
the distribution of centres is not only offset from the true value, but is noticeably skewed. 
Correcting each value by the corresponding expected bias \eqref{eq:centreBias}
gives a distribution consistent with the spread predicted by \eqref{eq:centreVariance}.

\section{Conclusions}

The recommended algorithm for optionally type-specific conic fitting with error estimation may be summarised as follows:---
\begin{enumerate}
\item From the given data $\{\vc x_i\}$, construct the design vectors $\{\vc D_i\}$.
\item \emph{Preliminary generic fit.} \label{algo:prelim}
\begin{enumerate}
\item
From $\{\vc D_i\}$, calculate the unweighted ($w_i=1$) scatter matrix $\mat S$. 
\item
Construct the self-normalising constraint matrix $\widetilde{\mat C}_\text{N}$
and calculate the reduced scatter matrix $\widetilde{\mat S}$.
\item
Solve the generalised eigenvalue problem \eqref{eq:reducedEigen} to find $\lambda_0$ and $\vc G_0$.
\item
Apply the curvature correction \eqref{eq:conicCorrection} to $\vc G_0$.
\end{enumerate}
\item \emph{Reweighting.} \label{algo:weight}
Using the elliptical coordinate system \eqref{eq:ellipticalCoords} generated by $\vc G_0$, 
for each data point $\vc x_i$:
\begin{enumerate}
\item
find the nearest point $\overline{\vc x}_i$ 
on the curve ${\vc G_0^\tp \vc D = 0}$;
\item
evaluate the gradient of $\vc D$ at $\overline{\vc x}_i$; and
\item
calculate a new weight $w_i$ for $\vc x_i$ from \eqref{eq:optimalWeight}.
\end{enumerate}
\item \emph{Weighted generic fit, with error estimate.}
\begin{enumerate}
\item
Repeat step \ref{algo:prelim}, but use the weights $\{w_i\}$ calculated in step \ref{algo:weight} 
and find all eigenvalues and corrected eigenvectors, not only $\vc G_0$.  
\item
Construct the generalised inverse $\mat Y_0$ as given by \eqref{eq:fullY}.
The covariance matrix for the generic fit is $\mat V_0=\sigma^2\mat Y_0/N$.   
\end{enumerate}
\item \emph{Type-specific fit.}
\begin{enumerate}
\item
Find the nearest parabolic solution $\overline{\vc G}$ using \eqref{eq:approxParabola}.  
If the unconstrained result $\vc G_0$ is already of the correct type, 
the first-order approximation will do; otherwise iterate \eqref{eq:approxParabola} as required.
\item
For an elliptic or hyperbolic fit, find the best point estimate of the conic parameters
$\expect{\vc G}$ from \eqref{eq:specificMean}.
\item 
For the parabolic case, the covariance is ${\overline {\mat V}=\sigma^2\overline{\mat Y}/N}$ 
where $\overline{\mat Y}$ is given by the generalised inverse of \eqref{eq:parabolicS}.  
For the elliptic or hyperbolic case, 
the probability distribution for $\vc G$ is the same as for the generic fit 
(Gaussian with mean $\vc G_0$, covariance $\mat V_0$), 
but truncated at the surface $\vc G=\overline{\vc G}$.
\end{enumerate}
\end{enumerate}

The preliminary generic fit is the same as recommended in \cite{collett2014ellipse},
and for practical purposes equivalent to the `semihyper-least-squares' method of \cite{kanatani2011hyper}.

This method contains exactly one reweighting.  
The effects of second or subsequent reweightings are much smaller,
and are unlikely to justify the extra computation.
Conversely, if no reweighting were performed, 
the generic covariance matrix would need to be calculated 
from \eqref{eq:covariance} instead of \eqref{eq:optimalCovariance};
the extra computational effort is similar to 
that required to calculate the weights, with less benefit. 
The reweighting is based on a generic fit, even if the final fit is type-specific;
we saw in Section \ref{sec:parabolic} that 
making the preliminary fit type-specific has no significant effect on the result.

It has been assumed throughout that 
the measurement noise is sufficiently small for the leading-order perturbative treatment to be valid. 
Specifically, $\sigma$ must be smaller than 
the smallest radius of curvature of the curve being fitted. 
For example, the curvature correction \eqref{eq:conicCorrection} is not accurate for larger noise.

Provided that this assumption is satisfied, the fitting method advocated in this paper produces 
an unbiased, minimal variance estimator of conic coefficients.
It has the advantages over previously published methods of 
giving confidence intervals rather than just point estimates,
and of allowing type-specific fitting while remaining unbiased.
It may be possible to make some of the individual steps more efficient 
(for example, by finding a more elegant method of extracting 
the mean or the covariance for the type-specific parabolic fit),
but not to improve significantly on the accuracy of the result in the small-noise regime. 

\bibliography{conicref}

\begin{thebibliography}{16}
\providecommand{\natexlab}[1]{#1}
\providecommand{\url}[1]{\texttt{#1}}
\expandafter\ifx\csname urlstyle\endcsname\relax
  \providecommand{\doi}[1]{doi: #1}\else
  \providecommand{\doi}{doi: \begingroup \urlstyle{rm}\Url}\fi

\bibitem[Bookstein(1979)]{bookstein1979fitting}
Fred~L. Bookstein.
\newblock Fitting conic sections to scattered data.
\newblock \emph{Computer Graphics and Image Processing}, 9\penalty0
  (1):\penalty0 56--71, 1979.

\bibitem[Sampson(1982)]{sampson1982fitting}
P.D. Sampson.
\newblock Fitting conic sections to very scattered data: An iterative
  refinement of the {Bookstein} algorithm.
\newblock \emph{Computer Graphics and Image Processing}, 18\penalty0
  (1):\penalty0 97--108, 1982.

\bibitem[Fitzgibbon et~al.(1999)Fitzgibbon, Pilu, and
  Fisher]{fitzgibbon1999direct}
Andrew Fitzgibbon, Maurizio Pilu, and Robert~B. Fisher.
\newblock Direct least square fitting of ellipses.
\newblock \emph{IEEE Transactions on Pattern Analysis and Machine
  Intelligence}, 21\penalty0 (5):\penalty0 476--480, 1999.

\bibitem[Halir and Flusser(1998)]{halir1998numerically}
R.~Halir and J.~Flusser.
\newblock {Numerically stable direct least squares fitting of ellipses}.
\newblock In \emph{6th International Conference in Central Europe on Computer
  Graphics and Visualization}, pages 125--132, February 1998.

\bibitem[Harker et~al.(2008)Harker, O'Leary, and
  Zsombor-Murray]{harker2008direct}
Matthew Harker, Paul O'Leary, and Paul Zsombor-Murray.
\newblock {Direct type-specific conic fitting and eigenvalue bias correction}.
\newblock \emph{Image and Vision Computing}, 26\penalty0 (3):\penalty0
  372--381, March 2008.

\bibitem[Kanatani(1994)]{kanatani1994statistical}
Kenichi Kanatani.
\newblock Statistical bias of conic fitting and renormalization.
\newblock \emph{IEEE Transactions on Pattern Analysis and Machine
  Intelligence}, 16\penalty0 (3):\penalty0 320--326, 1994.

\bibitem[Collett and Tee(2014)]{collett2014ellipse}
M.J. Collett and G.J. Tee.
\newblock Ellipse fitting for interferometry. {P}art 1: static methods.
\newblock \emph{Journal of the Optical Society of America A}, 31\penalty0
  (12):\penalty0 2573--2583, December 2014.

\bibitem[Kanatani and Rangarajan(2011)]{kanatani2011hyper}
Kenichi Kanatani and Prasanna Rangarajan.
\newblock Hyper least squares fitting of circles and ellipses.
\newblock \emph{Computational Statistics \& Data Analysis}, 55\penalty0
  (6):\penalty0 2197--2208, 2011.

\bibitem[Kanatani et~al.(2011)Kanatani, Rangarajan, Sugaya, and
  Niitsuma]{kanatani2011hyperls}
Kenichi Kanatani, Prasanna Rangarajan, Yasuyuki Sugaya, and Hirotaka Niitsuma.
\newblock {HyperLS} for parameter estimation in geometric fitting.
\newblock \emph{IPSJ Transactions on Computer Vision and Applications},
  3:\penalty0 80--94, 2011.

\bibitem[Kanatani et~al.(2015)Kanatani, Al-Sharadqah, Chernov, and
  Sugaya]{kanatani2015hyper}
Kenichi Kanatani, Ali Al-Sharadqah, Nikolai Chernov, and Yasuyuki Sugaya.
\newblock Hyper-renormalization: Non-minimization approach for geometric
  estimation.
\newblock \emph{Information and Media Technologies}, 10\penalty0 (1):\penalty0
  71--87, 2015.

\bibitem[Collett(2015)]{collett2015self-normalising}
M.J. Collett.
\newblock Self-normalising linear camera resection.
\newblock In \emph{IVCNZ'15}, volume~30, 2015.

\bibitem[Taubin(1991)]{taubin1991estimation}
Gabriel Taubin.
\newblock Estimation of planar curves, surfaces, and nonplanar space curves
  defined by implicit equations with applications to edge and range image
  segmentation.
\newblock \emph{IEEE Transactions on Pattern Analysis and Machine
  Intelligence}, 13\penalty0 (11):\penalty0 1115--1138, 1991.

\bibitem[Harker and O'Leary(2006)]{harker2006direct}
M.~Harker and P.~O'Leary.
\newblock Direct estimation of homogeneous vectors: An ill-solved problem in
  computer vision.
\newblock In \emph{ICVGIP'06}, pages 919--930. Springer, 2006.

\bibitem[Gander et~al.(1994)Gander, Golub, and
  Strebel]{gander1994least-squares}
W.~Gander, G.H. Golub, and R.~Strebel.
\newblock Least-squares fitting of circles and ellipses.
\newblock \emph{BIT Numerical Mathematics}, 34\penalty0 (4):\penalty0 558--578,
  1994.

\bibitem[Jaynes(2003)]{jaynes2003probability}
Edwin~T. Jaynes.
\newblock \emph{Probability theory: the logic of science}.
\newblock Cambridge University Press, 2003.

\bibitem[Jeffreys(1939)]{jeffreys1939theory}
H.~Jeffreys.
\newblock \emph{Theory of Probability}.
\newblock The Clarendon Press, Oxford, 1939.

\end{thebibliography}

\end{document}